\documentclass[sigconf,balance=false]{acmart}
\usepackage{popets}

\usepackage{tikz}
\usepackage{caption,subcaption}
\usepackage{graphicx}
\usepackage{array}
\usepackage{amsmath}
\usepackage{multirow}
\usepackage{diagbox}

\usepackage{graphicx} 
\usepackage{multirow}
\usepackage{array}
\usepackage{hyperref}

\def\ToolX{\textsc{SpinML}}

\setcopyright{popets}
\copyrightyear{YYYY}

\acmYear{YYYY}
\acmVolume{YYYY}
\acmNumber{X}
\acmDOI{XXXXXXX.XXXXXXX}
\acmISBN{}
\acmConference{Proceedings on Privacy Enhancing Technologies}
\begin{document}

\title[\ToolX: Customized Synthetic Data Generation for Private Training of Specialized ML Models]{\ToolX: 
 Customized Synthetic Data Generation for Private Training of Specialized ML Models}

\author{Jiang Zhang}
\orcid{1234-5678-9012}
\affiliation{%
  \institution{University of Southern California}
  \city{Los Angeles} 
  \state{California} 
  \country{USA} 
}
\email{jiangzha@usc.edu}

\author{Rohan Xavier Sequeira}
\affiliation{%
  \institution{University of Southern California}
  \city{Los Angeles} 
  \state{California} 
  \country{USA} 
}
\email{rsequeir@usc.edu}

\author{Konstantinos Psounis}
\affiliation{%
  \institution{University of Southern California}
  \city{Los Angeles} 
  \state{California} 
  \country{USA} 
}
\email{kpsounis@usc.edu}


\begin{abstract}
Specialized machine learning (ML) models tailored to users' needs and requests are increasingly being deployed on smart devices with cameras, to provide personalized intelligent services taking advantage of camera data.
However, two primary challenges hinder the training of such models: the lack of publicly available labeled data suitable for specialized tasks and the inaccessibility of labeled private data due to concerns about user privacy. 
To address these challenges, we propose a novel system \textbf{\ToolX}, where the server generates
customized \textbf{S}ynthetic image data to \textbf{P}rivately tra\textbf{IN} a specialized \textbf{ML} model tailored to the user request, with the usage of only a few sanitized reference images from the user. \ToolX~offers users fine-grained, object-level control over the reference images, which allows user to trade between the privacy and utility of the generated synthetic data according to their privacy preferences. Through experiments on three specialized model training tasks, we demonstrate that our proposed system can enhance the performance of specialized models without compromising users' privacy preferences.
\end{abstract}

\keywords{machine learning, synthetic data, privacy, utility}

\maketitle

\section{Introduction}
\label{sec:intro}
With the exponential growth of smart devices (e.g. smart speakers, smart monitors, smart watches), machine learning (ML) models are increasingly being deployed on such smart devices to provide intelligent services for users \cite{mahdavinejad2018machine,smartdevice2022}. A prominent trend in this evolution is the development of specialized ML models that are specifically tailored to the users' needs and requests, thereby enhancing the user experience across various applications \cite{atrey2021preserving}, including smart voice assistants \cite{edu2021skillvet}, wearable technologies \cite{bianchi2019iot}, specialized healthcare, \cite{sriramalakshmi2022modern}, etc.

Based on the availability and sensitivity of data and computation needs, specialized ML models can be trained either on the server or on local devices. For instance, in mobile applications, Google's Gboard \cite{hard2018federated} utilized Federated Learning \cite{mcmahan2017communication} to enhance the word prediction models by training them on user's devices thereby ensuring personal data privacy. Similarly, Apple utilizes on-device learning for Siri to improve speech recognition \cite{he2019streaming} without transmitting personal voice data to central servers. In retail, edge AI \cite{zhou2019edge} analyzes customer behavior in-store, enabling personalized recommendations without sending data to the cloud. These examples illustrate how the choice between server-based and local device training is determined by the need to balance performance optimization with privacy considerations.

Recent advancements in generative AI has further increased the need for personalized privacy-preserving ML models. For example,
"Apple Intelligence" \cite{apple2024intelligence} integrates generative AI models directly into iPhones, iPads, and Macs. The system takes advantage of large server-based models using Private Cloud Compute (PCC) which is designed to process complex AI tasks in the cloud while maintaining user privacy.
However, although Apple asserts that personal data sent to PCC isn't accessible to anyone other than the user, the inherent nature of cloud computing introduces risks, such as data breaches or unauthorized access. 
Emerging devices like Rabbit's r1 \cite{rabbit2024r1} and Humane's Ai Pin \cite{humane2024ai} further illustrate the trend toward integrating AI capabilities directly into consumer electronics. The Rabbit r1 is a pocket-sized AI companion that leverages a Large Action Model (LAM) to understand and execute user intentions with human-level reasoning. Humane's Ai Pin is a wearable device that acts as an intelligent, voice-powered companion, providing instant AI-powered knowledge and assistance.
However, while both systems emphasize privacy and security, they process user interactions through cloud-based Large Models and collect various types of user data, including precise geolocation, device information, and usage patterns, to provide personalized assistance. While these features enhance user experience, they also raise concerns about the extent of data collection and the potential for privacy leakage associated with cloud-based AI processing if the data isn't adequately protected. 
In summary, the aforementioned recent real-world scenarios clearly highlight the growing need for personalized intelligent services without compromising user's privacy. 

\begin{figure}[ht]
    \centering
    \includegraphics[width=\linewidth]{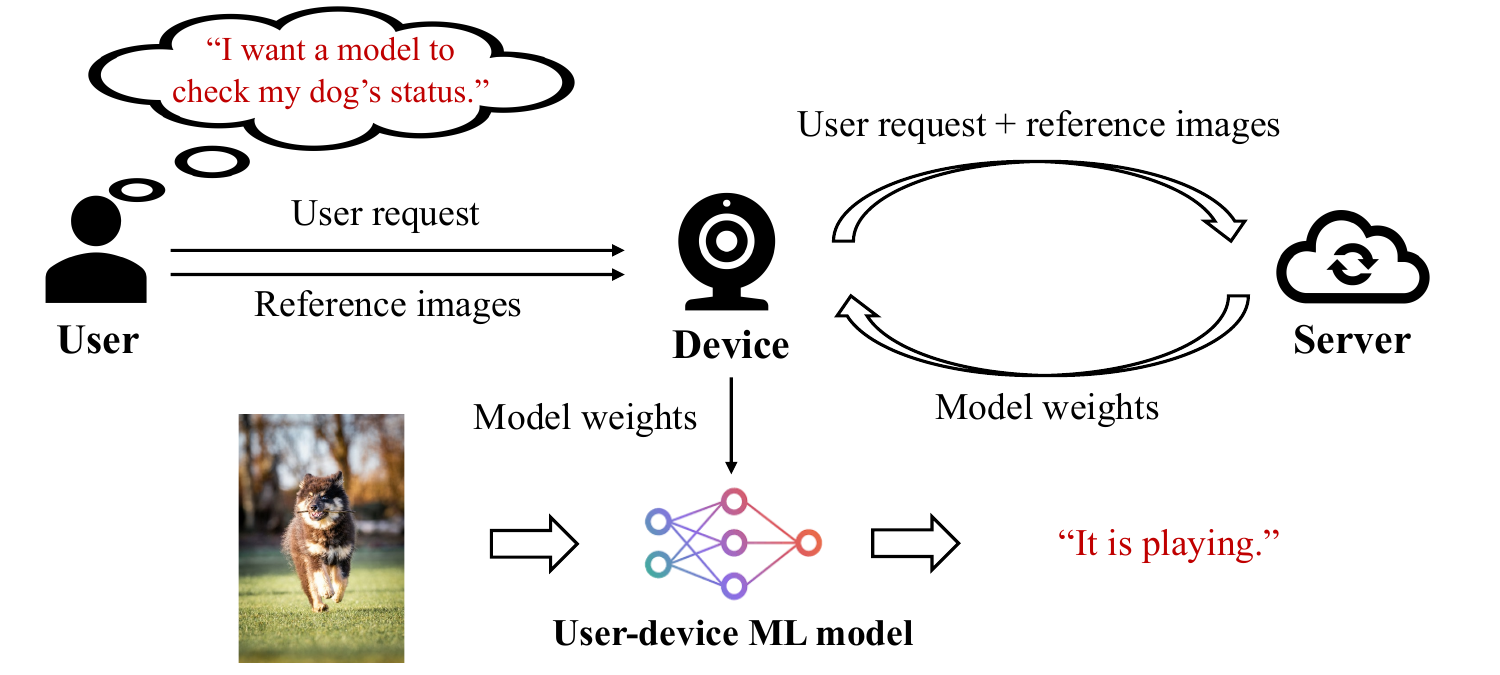}
    \caption{Problem statement. The user sends a request about the model they need and a few reference images. The server automatically train a model for the user.}
    \label{dm-fig:sys_overall}
    \vspace{-0.1in}
\end{figure}

In this paper, we focus on the scenario where the server trains specialized ML models tailored to unique user requests without accessing private user data (see Figure \ref{dm-fig:sys_overall}). The training process starts when the user sends a specialized ML model training request to the server  through the local device. Then, the server will automatically train an ML model tailored to user's request and deploy it on user's local device. Particularly, the user's local device is assumed to have no labeled dataset, but the user may be willing to share a few unlabeled data points with the server as references based on the user's privacy preference. 

There are three major challenges to training such specialized ML models that can satisfy the unique needs of users. First, public labeled data or models that fit the specialized tasks are often unavailable (e.g. training an image classifier for uncommon objects). Second, the server cannot have access to any labeled private data, due to privacy concerns and the lack of appropriate data annotation systems in the local devices. Third, due to the uniqueness of user request, other users' data cannot be leveraged to collaboratively train the specialized model (e.g. using federated learning \cite{kairouz2021advances}).

To handle these practical challenges, prior works have suggested utilizing high-quality synthetic data generated by large Diffusion Models (DMs) as training data \cite{he2022synthetic,yang2023diffusion}. While this approach provides a feasible alternative for generating large amounts of training data, the distribution of the generated synthetic data may significantly deviate from the user's private data, resulting in sub-optimal model performance. 
To better tailor the distribution of generated synthetic data, recent works have proposed methods for customizing large DMs. 
Given the prevalence of camera-equipped smart devices and the usefulness of camera data in multiple applications, these methods primarily focus on image data. 
Specifically, they involve fine-tuning the models on a set of reference images \cite{ruiz2023dreambooth} or incorporating conditional reference image features to the image generation process \cite{zhang2023adding}. 
Synthetic data generated by such customized DMs can significantly enhance the performance of specialized models \cite{fang2024data}. However, customizing DMs on reference images or image features can lead to privacy concerns, as sensitive information about users can leak from the reference images or image features shared with the server. The trade-offs between privacy and utility of customized synthetic data has not been well explored in prior research works. 

Therefore, in this paper, we propose \ToolX, a novel system to generate customized synthetic image data for specialized model training, which allows users to flexibly trade between the privacy and utility of generated synthetic data according to their privacy preferences. 
At a high level, our system provides users with fine-grained, object-level privacy control over the reference images shared with the server. This enables the selective removal of sensitive objects or features, while non-sensitive objects or features are retained and shared to maximize the utility of the customized synthetic data for training specialized models.

Specifically, \ToolX~ comprises three key components designed to enhance privacy and customization: 1) a light-weight object detection and segmentation module located on the local device, which partitions reference images into distinct image segments (e.g. target objects and background objects); 2) an image sanitizer, also on the local device, that removes sensitive features from each image segment according to users' privacy preferences; 3) a DM fine-tuning pipeline on the server side, designed to generate customized image segments and seamlessly merge them into the final synthetic images.

We evaluate \ToolX~ on three unique specialized model training tasks: pet status monitoring, human activity monitoring and non-popular object detection. Specifically, in the first task, the user request is to train a dog status classification model which can classify what the user's dog is doing at home. We consider the case where the user does not want to share the details of his/her home environment but may be willing to share some information about his/her dog to obtain a more accurate model. In the second task, the user request is to train a human activity classification model which classifies the activity of a senior person at home. We consider the case where the image of a senior person is treated as highly sensitive information while the home background may be shared for better model accuracy. In the third task, we assume that the users' main request would depend on a sub-request to train an accurate pill bottle detection model. We assume that the pill bottle contains highly sensitive information (e.g. label), while the user is willing to share some details about his/her home for better detection accuracy.  Our experimental results on these three case studies demonstrate the usability of \ToolX, which allows user to trade between privacy and utility flexibly in different situations. By using \ToolX, the performance of specialized models can be enhanced without compromising users’ privacy preferences.

\section{Preliminaries}
\subsection{Problem Statement}
\label{subsec:problem}
Our goal is to architect a system that achieves private training of specialized ML models for desirable user requests. On one hand, the system should have high utility, meaning that the specialized models trained via this system should achieve high inference accuracy on private users' data. To achieve the utility goal, the system needs to know enough information about the distribution of users' local data. On the other hand, the system should have high privacy, which means it does not breach user privacy when training the specialized model. To achieve the privacy goal, the system should not access any sensitive information about users' local data. At first glance, the privacy and utility goals appear to be at odds with each other. In this case, how can we have both high model accuracy and user privacy at the same time?

Table \ref{tab:motivation} provides some insights for addressing this challenging problem. Suppose that the users' local data can be split into sensitive parts and non-sensitive parts. For non-sensitive parts, they can always be kept in order to enhance model utility without hurting user privacy. For sensitive parts, if they do not affect the specialized model's accuracy, then they can be removed to maximize user privacy without hurting model utility. By contrast, if the sensitive parts affect specialized model's accuracy, it is unlikely to have both high accuracy and high utility. This says, the system has to trade between privacy and utility by sanitizing the data using different sanity levels.

With the above insights in mind, we design \ToolX, a system which provides users with fine-grained object-level sanity control over the reference images they share with the server. With \ToolX, users can flexibly trade between their data privacy and the utility of specialized models trained by the server based on their privacy preferences (see Section \ref{sec:approach} for design details).

\begin{table}[h]
\centering
\begin{tabular}{>{\centering\arraybackslash}p{0.9in}|>{\centering\arraybackslash}p{1.2in}|>{\centering\arraybackslash}p{0.8in}}
\toprule
\backslashbox{Utility}{Privacy} & Sensitive & Non-sensitive \\ \hline
Affect model utility  & Sanitize them (Either get privacy or get utility)  & Keep them (Maximize utility without hurting privacy)  \\ \hline
Not affect model utility & Remove them (Maximize privacy without hurting utility) & Do not matter (Simply keep them) \\ \bottomrule
\end{tabular}
\caption{Insights about how to trade between privacy and utility.}
\vspace{-0.1in}
\label{tab:motivation}
\end{table}

\subsection{Threat Model}
\noindent\textbf{User.} We assume that the user can request the server to build a specialized vision ML model tailored to a specific task and local, potentially resource constrained, device. The user request is a text prompt specifying the requirements for training a specialized model, which will subsequently be deployed locally on the users' local device. We assume that there is an absence of publicly available labeled datasets for training the specialized model. Additionally, we assume that the user might agree to share a limited number of sanitized reference images with the server to customize the generated synthetic data and thereby improve the accuracy of the specialized model. This consent can be acquired based on different levels of privacy preferences that will be available to the user.  

\noindent\textbf{Server.} We assume that the server has substantial computational resources to fine-tune and deploy large DMs for generating customized synthetic data. This data will then be used to train specialized local ML models (to be deployed on the user device) tailored to meet the users' specific requirements. Additionally, we assume that the server can be curious, implying that it may attempt to infer sensitive information from any data shared by the user. This includes both the text prompts detailing the users' requests and the sanitized reference images.

\subsection{Why Generic Models may not Work?}
\label{genericmodel}
There are three major reasons why generic models may not work well for personalized domains. 
First, generic models may have a fixed output format and fine-tuning is required to generate specialized output, such as predicting new classes. 
Second, they are trained on public data and they may not generalize to users’ private data. To address this, again fine-tuning may be required. 
Indeed,
our experimental results in Section \ref{sub-section-5.7:comparison-with-seem} demonstrate that large generic models perform poorly on personalized domains without additional fine-tuning.
Last, even if a large generic model (e.g. SEEM \cite{zou2023seem}) generalizes to a user’s private data in some circumstances, it is challenging to deploy it on mobile devices. 
In summary, there is a need to privately train personalized small models that can fit in local devices. 

\section{Design}
\label{sec:approach}
In this section, we present the design of \ToolX~ for private training of specialized ML models.
\begin{figure*}[!ht]
    \centering
    \includegraphics[width=0.8\linewidth]{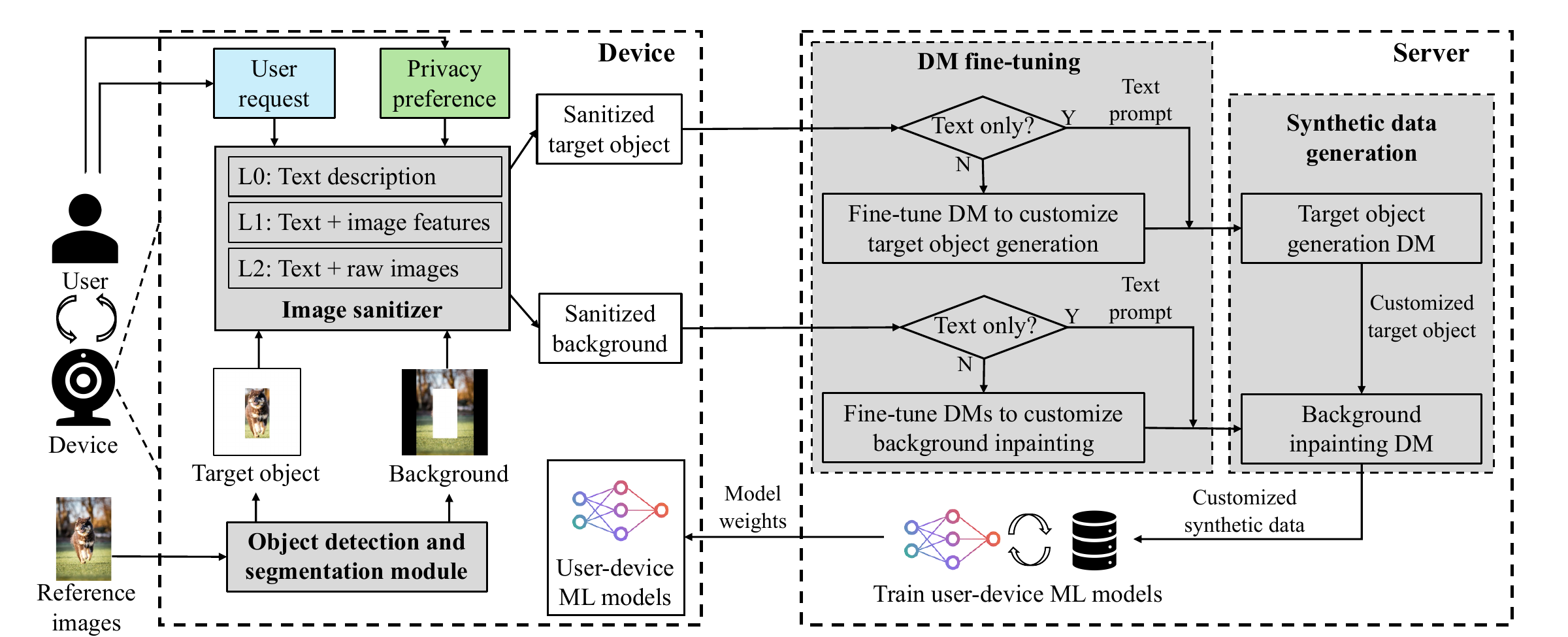}
    \vspace{-0.05in}
    \caption{System details of \ToolX. The user side consists of two modules: an object detection and segmentation module and an image sanitizer. The server side consists of three modules: a DM fine-tuning module, a synthetic data generation module, and a training module for a device-side ML model.}
    \label{dm-fig:sys_details}
    \vspace{-0.05in}
\end{figure*}
\subsection{System Overview}
Figure \ref{dm-fig:sys_details} demonstrates \ToolX, which contains multiple modules on both the user's device and the server. Specifically, on the device side, the system consists of: 1) an object detection and segmentation model for detecting and isolating target object from the background, and 2) an image sanitizer module for removing sensitive features in both the target object and the background (see Section \ref{dm-subsec:device}). The server-side system consists of three modules: 1) a DM fine-tuning pipeline to customize the target object and background generation of synthetic data, 2) a synthetic data generation module, and 3) a training module to train device-side ML models with the generated synthetic data (see Section \ref{dm-subsec:server} for details).

At a high level, \ToolX~ works as follows. First, the user specifies the training objective and requirements of the device-side ML model in a text request. The user has the option to share a few unlabeled private images with the server as reference training data. Suppose the user chooses to share a few reference images, then the image sanitizer generates sanitized images based on user's privacy preference and only sends the sanitized images to the server. Next, the server fine-tunes a DM \cite{rombach2022high} to generate customized synthetic data that satisfies user request and follows the distribution of sanitized reference images. Lastly, the server uses the customized synthetic data to train a device-side ML model and then sends the model weights back to depoly the model on the device.  We describe the design details below.

\subsection{Device-side System Design}
\label{dm-subsec:device}
\noindent\textbf{User request.} The user request comprises a list of key-value pairs specifying training requirements for the specialized ML model. As demonstrated in Table \ref{dm-tab:tab_user_req}, the user request contains four keys: target object, background, training object, label classes. The target object and background are used to instruct the server to generate images containing the specific types of target object and background. For example, if the user wishes to have a specialized ML model to monitor their dog's status in their room, the target object and the background would be specified as ``dog'' and ``bedroom'' respectively, such that the server can generate images of a dog in a room. Moreover, the training object is used to indicate the functionality of the specialized ML model, and the label classes specify the categories of the generated images. For instance, in the scenario involving monitoring the dog's status, the label classes can include ``eating'', ``sitting'', ``sleeping'', ``playing'', which will be used by the server to generate dog images with specified status.

\begin{table}[h]
\centering
\begin{tabular}{p{1.1in}<{\centering}|p{1.8in}<{\centering}}
\toprule
Key & Value example \\ \hline
Target object & Dog \\
Background & Bedroom \\
Training objective & A ML model detects my dog's status \\ 
Label classes & Eating, sitting, sleeping, playing \\
\bottomrule
\end{tabular}
\caption{An example of user request, where the goal is to train a specialized ML model to monitoring the dog's status in the user's room.}
\vspace{-0.1in}
\label{dm-tab:tab_user_req}
\end{table}

\noindent\textbf{Object detection and segmentation module.} The purpose of this module is to identify and separate the objects in the reference images, which enables fine-grained privacy control for each object based on the user's individual privacy preferences. 
In this work, we group the objects within an image into two categories: target objects and background objects. The target objects, which are specified by the user in their request, are the primary focus of the model's training. As an example, if the user request is to train a ML model to monitor the dog's status, the target object will be dog.

The object detection and segmentation module is built based on YOLO-v8, an off-the-shelf object detection and segmentation model for end devices \cite{Yolov8Ultralytics}.
For each reference image, the module runs the object detection and segmentation model to detect and segment a list of objects. Based on the target object defined in the user request, this module splits the images into two distinct segments: one containing only the target object (the target image) and another containing all remaining objects (the background image). This module allows users to control the privacy of different image segments according to their specific needs.

\noindent\textbf{Image sanitizer.} This module is designed to generate text descriptions of the input image segments and remove sensitive features from them like the target object or background. 
The text description for the target object is simply the value of the target object specified in the user request.
Similarly, the text description for the background is simply the value of the background specified in the user request.
%

Crucially, the image sanitizer provides three sanitization schemes to remove sensitive features from the input, which we denote by $L_0$, $L_1$ and $L_2$ respectively. We describe the details of these schemes below.
\begin{enumerate}
    \item $L_0$: This scheme only generates a text description of the input image segment (either target object or background) and sends it to the server. As an example, if the input image segment contains a dog, the output text of the image sanitizer under L0 scheme will be ``dog''. Note that the text description of the target object and background can be derived from the user request (see Table \ref{dm-tab:tab_user_req} for example). When the input image segment is highly sensitive, $L_0$ should be used to maximize  privacy.
    \item $L_1$: This scheme extracts non-sensitive image features from the input image segment, and sends both these features and the text description of the image to the server. Examples of extracted image features include canny edge, object skeleton, layout box, etc., which can be used for conditional synthetic image generation \cite{zhang2023adding}. $L_1$ is suitable for use when the input image segment features are considered to be non-sensitive for users. 
    \item $L_2$: This scheme sends the raw input image segment along with the text description of the image segment to the server. When the input image segment is non-sensitive, using $L_2$ is recommended to maximize the utility of the sanitized output. 
    \vspace{-0.1in}
\end{enumerate}
For each reference image, after the object detection and segmentation module has isolated the target object and background, the image sanitizer processes them in parallel to produce sanitized versions of both the target object and background. Note that different sanitization schemes may be applied to the target object and background as needed. When $L_0$ scheme is applied to the target object or background, the sanitized target object or background is represented by text descriptions only. When $L_1$ or $L_2$ scheme is applied to the target object or background, the sanitized target object or background includes both text and sanitized reference images, which comprise either image features or raw images, depending on the scheme applied.

\noindent\textbf{User's privacy preference.} We define the user's privacy preference as ($L_i^t$, $L_i^b$), $i\in\{0,1,2\}$, where $L_i^t$ indicates that $L_i$ scheme is used by the image sanitizer for the target object and $L_i^b$ indicates that $L_i$ scheme is used by the image sanitizer for the background. This dual parameterization allows the user to choose a different sanitization level for the target object and the background. Note that \ToolX~ can be generalized to multi-object scenarios where each object has different sanity levels (see Section \ref{sec:discussion} for a detailed discussion).

\subsection{Server-side System Design}
\label{dm-subsec:server}
\noindent\textbf{DM fine-tuning module.} This module is designed to fine-tune the DM in order to generate both the customized target object and the background when sanitized reference images are shared. Upon receiving the sanitized target object and background from the user device, it first checks whether the sanitized target object contains any image data. If the sanitized target object contains text only (i.e. $L_0$ is used), the server does not fine-tune the target object generation DM. Instead, it prompts a pre-trained DM using the text to generate the target object. In contrast, if the server receives sanitized reference images (i.e. $L_1$ and $L_2$ are used), it fine-tunes the target object generation DM using these images, in order to produce synthetic target objects whose distribution is similar to the distribution of the sanitized reference images. Specifically, if the sanitized reference images contain image features only (i.e. $L_1$ scheme is used), this module initially employs ControlNet \cite{zhang2023adding} to generate a set of new reference images conditional on these image features. Following this, the module fine-tunes a pre-trained DM on these new reference images using the DreamBooth algorithm \cite{ruiz2023dreambooth}. If the sanitized reference images are generated by $L_2$ scheme (which includes raw images), this module directly fine-tunes the pre-trained DM on these reference images through the DreamBooth algorithm.

Similarly, if the server receives sanitized reference images for the background, it will fine-tune the DM to generate synthetic background images that are aligned with the reference background images. Otherwise, it uses the pre-trained DM without any fine-tuning to generate synthetic background images.

\noindent\textbf{Synthetic data generation module.} This module operates in three sequential steps to produce synthetic data. Initially, it utilizes the customized target object generation DM to create the target object using a set of text prompts. Note that the text prompts for generating the target object consist of all possible combinations of the text description of the target object and each label class. For example, in the task of monitoring a dog's status (see Table \ref{dm-tab:tab_user_req}), the set of text prompts would include: ``a dog is eating," ``a dog is sitting," ``a dog is sleeping," and ``a dog is playing". \footnote{Note that image features are used to fine-tune the DM as discussed above.} 
%
Next, the synthetic data generation module generates a random background mask for the target object. Lastly, it employs the customized background generation DM to fill the background mask and seamlessly integrate the background with the generated target object. \footnote{Note that the text prompts for generating the background would simply be the text description of the background shared by the image sanitizer.} Notably, during our experiments, we merge the weights of the fine-tuned background generation DM with those of a pre-trained inpainting DM to create a customized background inpainting DM specifically tailored for generating the background.


\noindent\textbf{Specialized ML model training.} After the synthetic data generation module produces a set of synthetic data, this module uses the synthetic data to train a specialized ML model via supervised learning. 
Specifically, for image classification tasks where the text prompt explicitly specifies the label of the generated synthetic images, the label can be directly extracted from the text prompt. For instance, if the text prompt used for target object generation is ``a dog is running'', the label for this image would be ``running''. 
For object detection tasks where the text prompt does not explicitly specify label information, the server automatically obtains the label
during the synthetic data generation process, as the label is merely the location of the object, see Section \ref{dm-subsec:train_and_test} for a more detailed discussion.

\subsection{Privacy Measurement}
\label{dm-subsec:priv-measure}
As described in Section \ref{dm-subsec:device}, the image sanitizer will share the text description of the image segments and the sanitized image segments with the server for customized synthetic data generation. Therefore, the user's private information may leak through both the text description of image segments and the sanitized image segments.
To quantify the privacy leakage from the image sanitizer, we measure the average Mutual Information (MI) between reference image segments and the corresponding sanitized reference image segments generated by the sanitizer. 
Because MI does not measure semantic information leakage, we also use Semantic Embedding Similarity (SIM) as a second metric to measure the semantic information leakage between private and generated synthetic images. We discuss in further detail these two privacy metrics below.
%

\noindent\textbf{MI} is a formal metric based on information theory to quantify on-average privacy leakage between each reference image and the corresponding sanitized reference image generated by the image sanitizer, see \cite{cover1999elements} for a formal definition of the mutual information between two random variables and \cite{russakoff2004image} for how to compute the mutual information between two images. Formally, suppose that the user is willing to share a set of $N$ reference images denoted as $X=\{x_{1},...,x_{N}\}$). With the object detection and segmentation module, each reference image $x_{i}$ will be split into a target object image $x^t_{i}$ and a background image $x^b_{i}$. We define $X^t=\{x^t_{1},...,x^t_{N}\}$ as the set of reference target object images and $X^b=\{x^b_{1},...,x^b_{N}\}$ as the set of reference background images. After sending the reference target object and background images to the image sanitizers, a set of $N$ sanitized reference target object images $Y^t=\{y^t_{1},...,y^t_{N}\}$ and a set of $N$ sanitized reference target object images $Y^b=\{y^b_{1},...,y^b_{N}\}$ will be generated and shared with the server. Then, we formally define our MI-based privacy metric as follows:
\begin{equation}
    MI(X^{j}, \{Y^t,Y^b\}) = \frac{1}{N}\sum_{i=1}^{N}\frac{I(x^j_i,\{y^t_i,y^b_i\})}{I(x^j_i,x^j_i)},
\end{equation}
where $j\in\{t,b\}$ indicates target object or background, $I(A,B)$ denotes the mutual information between image $A$ and $B$, $I(x^j_i,\{y^t_i,y^b_i\})$ quantifies how much information the sanitized reference target object and background image $\{y^t_i,y^b_i\}$ together leaks about the raw reference target object image ($x^t_i$) or background image ($x^b_i$), and $I(x_i,x_i)$ quantifies how much information $x_i$ leaks about itself, i.e., the entropy of raw reference image $x_i$. 

$MI(X^{t}, \{Y^t,Y^b\})$ and $MI(X^{b}, \{Y^t,Y^b\})$ measure the average percentage of information leakage about the target object and background, respectively, in the user's raw reference images given the output of the image sanitizer $\{Y^t,Y^b\}$. Higher $MI(X^{j}, \{Y^t,Y^b\})~(j\in\{t,b\})$ means more privacy leakage from the sanitized reference images output by the image sanitizer. 

Note that $MI(X^{j}, \{Y^t,Y^b\})$ $(j\in\{t,b\})$ does not consider the potential information leakage via the text description as the text description is deterministic (e.g. when the target object is a dog, the image sanitizer will always send ``dog" as the text description to the server), and thus the MI between the text description and the raw reference images would be zero. \footnote{Note that, in general, to formally calculate the MI between a text and an image, we could map both into a common embedding space, and then estimate the MI between the two embeddings.} For instance, when $L_0$ is applied to the user's dog images (target object), the server will not be able to learn any information about how the user's dog looks like. However, the server will know that the user has a dog, which may be considered as sensitive for certain users. To measure the end-to-end privacy of the whole \ToolX~system, we define the following semantic-based privacy metric.

\noindent\textbf{SIM} is defined as the Semantic Embedding Similarity (SIM) between the user's private images and synthetic images generated by the server. Specifically, we first employ a large pre-trained vision embedding model to generate the semantic embedding for each private and synthetic image. \footnote{Note that we use Blip2 \cite{li2023blip} to generate the semantic embedding of each image.} We then calculate the average cosine similarity between each pair of private image embedding and synthetic image embedding. Note that we measure the SIM of the target object and background separately. Specifically, suppose that $P^t$ and $P^b$ are two image sets, containing target object images and background images in the user's private dataset respectively, and $Q^t$ and $Q^b$ are two image sets, containing target object images and background images in the synthetic dataset generated by the server respectively. Formally, the SIM metric is defined as:
\begin{equation}
    SIM(P^j,Q^j) = E_{p^j_i,q^j_i}[Cos(Emb(p^j_i), Emb(q^j_i))],
\end{equation}
where $j\in\{t,b\}$ indicates the target object or the background, $Cos(x,y)$ measures the cosine similarity between vector $x$ and $y$, $p^j_i\in P^j$ is an image instance from the user's private dataset, $q^j_i\in Q^j$ is an image instance from the synthetic dataset generated by the server, and $Emb(x)$ denotes the semantic embedding of image $x$. 

Note that $SIM(P^t,Q^t)$ measures the average similarity between the synthetic target object generated by the server and the raw target object in the user's private dataset, and $SIM(P^b,Q^b)$ measures the average similarity between the synthetic background generated by the server and the raw background in the user's private dataset. Higher $SIM(P^j,Q^j)~j\in\{t,b\}$ indicates more privacy leakage, as it suggests that the synthetic target object, or background, generated by the server is more similar to the one in the user's private dataset. If the synthetic images generated by the server were the same as the user's private images, $SIM(P^j,Q^j)$ would be equal to one. 

It is worth mentioning that while the text prompt contains semantic information based on which the synthetic image is generated, the associated leakage from the prompt cannot be measured by comparing original and synthetic images. To measure the semantic information leakage between text prompts and private images directly we use a pre-trained vision-language model proposed in \cite{zeng2023clip2} to map both text prompts and private images into the same embedding space, and then compute the SIM metric between the text prompt embedding and the private image embedding as a measurement of semantic information leakage from text prompts. We report results of this metric in Section \ref{subsec:privacy}.

\noindent
\textbf{Remark:}
The privacy metrics we define here measure on-average privacy, rather than worst-case privacy guarantees such like differential privacy (DP) \cite{dwork2014algorithmic}. There are two reasons for not using DP as a privacy metric in our context. First, it would be very challenging to estimate the worst-case privacy bound ($\epsilon$ value) that might be provided when sharing, say, image features, since there is no direct DP noise addition. Second, achieving worst-case privacy guarantees can significantly degrade the utility of the sanitized reference images. Indeed, as we show in 
Section \ref{subsec:obfuscation}, adding noise of variable levels performs worse than using the 
$L_0$ scheme (i.e. not sharing any image data/features) with respect to both 
accuracy and privacy (when using MI as the privacy metric), and, to achieve any meaningful $\epsilon$ to offer worst case privacy, one would have to add so much noise that the model accuracy would be prohibitively low for any practical usage. 

\section{Experimental setup}
\label{dm-sec:setup}
\subsection{Real-world Tasks and Datasets}
\label{dm-subsec:real-data}
\noindent\textbf{Pet status monitoring.} This task involves training a mobile ML model to monitor the status of pets for the user. We create a user dataset containing husky dogs, with behaviors categorized into four statuses: playing, eating, sitting, sleeping. We refer to this dataset as husky dataset, and use it to evaluate the accuracy of the end-device ML model trained by the server.

\noindent\textbf{Human activity monitoring.} This task focuses on training an end-device ML model to monitor the daily activities of senior individuals within a home environment. We utilize a subset of the Toyota Smarthome dataset \cite{das2019toyota}, which contains 16.1K video clips with 31 activity classes performed by 18 senior people in a large house with 7 cameras. Specifically, we sample the image frames of a single senior person engaging in four activities: eating, drinking, walking, reading. We refer to the sampled data as the human dataset.

\noindent\textbf{Non-popular object detection.} This task involves fine-tuning an object detection model to detect non-popular objects. As a case study, we consider the task of detecting a (medicine) pill bottle in a bedroom environment, and we create an image dataset consisting of a pill bottle in a bedroom to evaluate the detection accuracy of a corresponding end-device ML model. This dataset is named as the bottle dataset.

\subsection{Models and Synthetic Datasets}
\label{dm-subsec:syn-data}
We employ the YOLO-v8 segmentation model \cite{Yolov8Ultralytics} in the object detection and segmentation module to separate the target object from the background. For the generation process, we utilize Stable-Diffusion-v1.5 \cite{rombach2022high} as the pre-trained Diffusion Model (DM) for both the target object and the background generation in our experiments. We describe how we generate customized synthetic datasets for each task in detail below.

\noindent\textbf{Pet status monitoring.} In this task, we consider the background to be sensitive and hence apply $L_0$ sanitization scheme to it. For the target object (i.e. the husky dog in this case), we assume that the user may have different privacy preferences. Therefore, we apply all three different sanitization schemes for the target object ($L_0$, $L_1$ and $L_2$). This results in three different user privacy preferences: ($L_0^t$, $L_0^b$), ($L_1^t$, $L_0^b$), ($L_2^t$, $L_0^b$). For each user privacy preference, we generate 1,600 synthetic images (400 for each image class), and use them to train a dog status classifier. Note that for ($L_0^t$, $L_0^b$), we use the pre-trained DM without fine-tuning to generate synthetic images. For ($L_1^t$, $L_0^b$), we select the canny edge as the image feature, and the image sanitizer extracts the canny edge of the target object and shares it with the server. The server then uses ControlNet \cite{zhang2023adding}, which takes both the canny edge and the text description of the target object (i.e. ``a dog") as input, to generate a set of synthetic reference dog images with the same canny edge. The synthetic reference dog images are further used to fine-tune DM for customized target object generation. For ($L_2^t$, $L_0^b$), the image sanitizer directly shares the target object (i.e. the husky dog) in these reference images with the server, and the server will fine-tune the DM to generate images containing husky dogs which are similar to the one shared by the image sanitizer.

\noindent\textbf{Human activity monitoring.} In this task, we consider the target object (i.e. the senior person) as highly sensitive and apply the $L_0$ and $L_1$ sanitization schemes to it. For the background (i.e. the home environment), we assume that the user may have different privacy preferences and thus we consider all three sanitization schemes for it (ranging from $L_0$ to $L_2$). This leads to six different user privacy preferences: ($L_0^t$, $L_0^b$), ($L_0^t$, $L_1^b$), ($L_0^t$, $L_2^b$), ($L_1^t$, $L_0^b$), ($L_1^t$, $L_1^b$), ($L_1^t$, $L_2^b$). For each user privacy preference, we generate 1,600 synthetic images (400 for each image class), which are used to train a human activity classifier. Note that for the $L_1^b$ scheme, we select the canny edge as the image feature, and the image sanitizer shares the canny edges of the background with the server. For the $L_1^t$ scheme, we select the human pose as the image feature, and the image sanitizer shares the human pose of the target object with the server. 

\noindent\textbf{Non-popular object detection.} In this task, the pill bottle, i.e. the target object,  is considered to have highly sensitive content, e.g. a medical pill label, and hence we apply either the $L_0$ or the $L_1$ scheme to it. For the background, we assume that the user may be willing to share the raw image (i.e. the bedroom), and thus we offer all three sanitization schemes for it (ranging from $L_0$ to $L_2$). In total, we implement six user privacy preferences: ($L_0^t$, $L_0^b$), ($L_0^t$, $L_1^b$), ($L_0^t$, $L_2^b$), ($L_1^t$, $L_0^b$), ($L_1^t$, $L_1^b$), ($L_1^t$, $L_2^b$). For each user privacy preference, we generate 1,600 synthetic images where the pill bottles are randomly placed in the image and the location of each pill bottle is the label. Note that we select the canny edge as the image feature for the $L_1$ scheme for both the target object and the background.

\subsection{Training and Testing}
\label{dm-subsec:train_and_test}
For the pet status and human activity monitoring tasks, we use MobileNet-v2 \cite{sandler2018mobilenetv2} as the backbone, and then add a linear layer followed by a softmax layer as our specialized end-device ML models. Note that we use the classification accuracy as the metric to measure the performance of the specialized models. For the non-popular object detection task, we use the pre-trained YOLO-v8 detection model \cite{Yolov8Ultralytics} as the specialized model. To measure the performance of this specialized model, we use mAP50 (mean average precision calculated at an intersection over union (IoU) threshold of 0.50) as the metric, which is commonly used in object detection.

During the training process, we use 80\% of the synthetic data for training and the remaining 20\% of the synthetic data for validation. For each user privacy preference in each task, we train each specialized model for 5 epochs and select the model with the highest validation accuracy as our final model. During the testing phase, we test the accuracy of these specialized models on real-world data (see Section \ref{dm-subsec:real-data} for more details).

\section{Evaluation}
\subsection{Utility}
\label{subsec:utility}
We first evaluate the utility of customized synthetic data with various user privacy preferences, using the performance of specialized ML models as the metric (i.e. accuracy for the husky and human datasets, and mAP for the bottle dataset). We report the model performance results on the three tasks in Table \ref{dm-tab:utility}. Note that we report the model performance on both the validation sets of the generated synthetic data and on the users' private data on their local devices which represents the ``real world" dataset. Unless otherwise stated, we focus on the real-world dataset results since they are the one that matter in practice.

First, we observe that using the $L_2$ scheme for target object or background sanitization can significantly improve the model accuracy trained with the customized synthetic data on all three real-world datasets. This is expected since $L_2$ sends the parts of the raw reference images (target object or background) to the server, which enables the fine-tuned DMs to generate synthetic images that closely resemble these raw reference images.

Next, we observe that using the $L_1$ scheme for target object or background sanitization may not always lead to better model accuracy. Since $L_1$ sends features (canny edge for husky and bottle datasets and human pose for human dataset) of the reference images to the server, whether these image features can help to generate better synthetic images depends on whether these features are important to the model utility. For instance, prior works have shown that canny edges are important features for object detection \cite{kim2020study}; Indeed, during our experiments on the bottle dataset, we consistently observe that using the $L_1$ scheme effectively boosts the performance of specialized models. Moreover, whether the customized synthetic data can improve the model accuracy depends on whether the shared features can be leveraged to fine-tune the DM properly. As an example, for the husky dataset, without knowing the label (i.e. the status of the dog) of each canny edge image, the server may not know the proper correlation between the canny edge and the status of the dog, and thus the DM may not be properly fine-tuned (see Section \ref{subsec:putrade} for a detailed discussion).

Last, we find that specialized models trained on synthetic data may overfit the training data, and hence they may fail to generalize to the real-world testing data well. As shown in Table \ref{dm-tab:utility}, the model accuracy on synthetic validation data (see Section \ref{dm-subsec:syn-data}) is higher than that in the real-world testing data for the Husky and Human use cases, since synthetic validation data has the same distribution with the synthetic training data, while the real-world testing data does not. To mitigate the overfitting issue, we sample the training data from synthetic data generated by different schemes, and we observe this approach may help improve the model accuracy. For example, on the husky dataset (see Table \ref{dm-tab:utility1}), combining the synthetic data generated with $L_0$ and $L_2$ for the target object increases model accuracy on real-world data by 8.34\%. Last, note that in the pill bottle use case, the pill bottle generated by the diffusion model is less distinguishable under the ($L_2^t$, $L_2^b$) scheme as compared to the other schemes, resulting in inconsistent accuracy results. We believe this is due to the relatively small diffusion model that we use due to resource limitations, see Section \ref{sec:discussion} for a more detailed discussion. 

Note that on the human dataset the specialized model trained on synthetic data has low accuracy (below 50\%) as compared to the models trained for the other two case studies / datasets. This is mainly because for the human activity monitoring task, video input rather than just images may be needed for the model to classify such human activities accurately, see prior work \cite{dai2022toyota} and a discussion about the reasons for this limitation in our study in Section \ref{sec:discussion}.

\begin{table}[t]
\centering
\small
\begin{subtable}{0.45\textwidth}
\begin{tabular}{p{1.2in}<{\centering}p{0.8in}<{\centering}p{0.8in}<{\centering}}
\hline
\multirow{2}{*}{Privacy preference} &  \multicolumn{2}{c}{Accuracy}\\\cline{2-3}
 & Synthetic data & Real-word data \\\hline
($L_0^t$, $L_0^b$) & 87.88\% & 63.46\% \\
($L_1^t$, $L_0^b$) & 83.45\% & 57.05\% \\
($L_2^t$, $L_0^b$) & 88.74\% & 64.74\% \\
($L_2^t$, $L_2^b$) & 98.33\% & 65.38\% \\
($L_0^t$, $L_0^b$) + ($L_1^t$, $L_0^b$) & 81.82\% & 62.82\% \\
($L_0^t$, $L_0^b$) + ($L_2^t$, $L_0^b$) & 83.10\% & \textbf{73.08\%}\\
\hline
\end{tabular}
\caption{Husky dataset which contains images of husky dogs exhibiting four statuses: playing, running, sitting, and sleeping. 
}
\label{dm-tab:utility1}
\end{subtable}
\begin{subtable}{0.45\textwidth}
\begin{tabular}{p{1.2in}<{\centering}p{0.8in}<{\centering}p{0.8in}<{\centering}}
\hline
\multirow{2}{*}{Privacy preference} &  \multicolumn{2}{c}{Accuracy}\\\cline{2-3}
 & Synthetic data & Real-word data \\\hline
($L_0^t$, $L_0^b$) & 74.26\% & 41.20\% \\
($L_0^t$, $L_1^b$) & 66.04\% & 44.05\% \\
($L_0^t$, $L_2^b$) & 61.48\% & 49.42\% \\
($L_1^t$, $L_0^b$) & 77.50\% & 41.54\% \\
($L_1^t$, $L_1^b$) & 74.79\% & 45.23\% \\
($L_1^t$, $L_2^b$) & 81.46\% & 42.21\% \\
($L_2^t$, $L_2^b$) & 86.25\% & 44.22\% \\
($L_0^t$, $L_2^b$) + ($L_1^t$, $L_2^b$) & 70.21\% & 45.39\% \\
\hline
\end{tabular}
\caption{Human dataset which contains images of a single senior person in a home environment engaged in four activities: eating, drinking, walking, and reading. 
}
\label{dm-tab:utility2}
\end{subtable}
\begin{subtable}{0.45\textwidth}
\begin{tabular}{p{1.2in}<{\centering}p{0.8in}<{\centering}p{0.8in}<{\centering}}
\hline
\multirow{2}{*}{Privacy preference} &  \multicolumn{2}{c}{Accuracy (mAP50)}\\\cline{2-3}
 & Synthetic data & Real-word data \\\hline
($L_0^t$, $L_0^b$) & 86.75\% & 20.85\% \\
($L_0^t$, $L_1^b$) & 91.93\% & 57.52\% \\
($L_0^t$, $L_2^b$) & 90.94\% & 87.96\% \\
($L_1^t$, $L_0^b$) & 76.01\% & 93.59\% \\
($L_1^t$, $L_1^b$) & 91.86\% & 95.88\% \\
($L_1^t$, $L_2^b$) & 73.42\% & 97.94\% \\
($L_2^t$, $L_2^b$) & 64.98\% & 99.21\% \\
($L_0^t$, $L_2^b$) + ($L_1^t$, $L_2^b$) & 78.90\% & 98.32\% \\
\hline
\end{tabular}
\caption{Bottle dataset which contains images of a medicine pill bottle located in a bedroom.
}
\label{dm-tab:utility3}
\end{subtable}
\caption{Utility evaluation results. Note that we use accuracy as the utility metric for the husky and human datasets, and mAP50 (mean average precision calculated at an intersection over union (IoU) threshold of 0.50) as the utility metric for the bottle dataset.}
\label{dm-tab:utility}
\vspace{-0.1in}
\end{table}

\subsection{Privacy}
\label{subsec:privacy}
Next, we evaluate the privacy of customized synthetic data with various user privacy preferences using the two privacy metrics mentioned in Section \ref{dm-subsec:priv-measure}. We report the results in Table \ref{dm-tab:privacy}. Note that MI measures the average mutual information between the user's raw reference images and the sanitized reference images shared with the server. Higher MI value indicates that sanitized reference images shared with the server leak more privacy information about the user's raw reference images. SIM quantifies the semantic embedding similarity between the user's private image and the generated synthetic images or the text prompts used for generating synthetic images. A higher SIM value indicates that the synthetic images generated by the server are more similar to the user's private images, thereby leaking more private information. Moreover, for each of these two privacy metrics, we report the privacy leakage w.r.t. the target object and background separately to demonstrate where the privacy leakage takes place.

On the Husky and Human datasets, we observe that both the synthetic data generated by $L_1$ and $L_2$ schemes exhibit higher SIM scores compared with the data generated by $L_0$, as expected (see Table \ref{dm-tab:privacy1} and Table \ref{dm-tab:privacy2}). This indicates that the synthetic data generated by $L_1$ and $L_2$ are more similar to the user’s private data compared to $L_0$, causing more privacy leakage. Specifically, for the Husky dataset, the SIM scores for the target object and background both increase as we move from $L_0$ to $L_2$. For the Human dataset, a similar trend is observed, with the SIM scores for both the target object and background increasing from $L_0$ to $L_2$.

In terms of MI, for the Husky dataset, there is a significant increase in MI for both the target object and background as we move from $L_0$ to $L_2$, indicating a higher degree of privacy leakage, as expected (see Table \ref{dm-tab:privacy1}). For the Human dataset, the MI for both the target object and background also increases from $L_0$ to $L_2$, though the increase is more moderate compared to the Husky dataset (see Table \ref{dm-tab:privacy2}). 
On both datasets, ($L_2^t$, $L_2^b$) privacy preference leads to the highest MI and SIM scores.

On the Bottle dataset (see Table \ref{dm-tab:privacy3}), we observe a similar trend with respect to SIM scores. Despite the increased SIM scores observed with $L_1$ and $L_2$ schemes, it is worth noting that these increases are relatively modest in this case study, indicating only a moderate enhancement in the similarity of the synthetic data to the user’s private data. In terms of MI, the MI for the target object and background shows a significant increase from $L_0$ to $L_2$ as before, indicating a higher degree of privacy leakage, as expected. Consistent with results on other two datasets, ($L_2^t$, $L_2^b$) privacy preference leads to both the highest MI and SIM scores.
%

Last, in Table \ref{dm-tab:privacy4}, we report the SIM score of text prompts and private images to measure the semantic information leakage from prompts. Note that we use as baseline the SIM score of an empty text prompt. As demonstrated in Table \ref{dm-tab:privacy4}, the SIM score of text prompts is close to the baseline. This indicates that the semantic similarity between private images and text prompts is similar to that between private images and an empty prompt, implying that the information leakage from text prompts is small.

\begin{table}[!h]
\centering
\small
\begin{subtable}{0.48\textwidth}
\begin{tabular}{p{0.45in}<{\centering}p{0.6in}<{\centering}p{0.5in}<{\centering}p{0.6in}<{\centering}p{0.5in}<{\centering}}
\toprule
\multirow{2}{*}{\shortstack{Privacy \\ preference}} &  \multicolumn{2}{c}{MI} & \multicolumn{2}{c}{SIM } \\\cline{2-5}
 & Target Object & Background & Target Object & Background \\\hline
($L_0^t$, $L_0^b$) & 0.00\% & 0.00\% & 0.62 & 0.55 \\
($L_1^t$, $L_0^b$) & 0.89\% & 0.70\% & 0.72 & 0.57 \\
($L_2^t$, $L_0^b$) & 39.83\% & 11.31\% & 0.72 &  0.57 \\
($L_2^t$, $L_2^b$) & 39.83\% & 71.48\%   & 0.72 &  0.59 \\
\bottomrule
\end{tabular}
\caption{Husky dataset. Note that the target object in this dataset is husky dog, and the background can be both indoor and outdoor environment. We assume that the user mainly cares about the background privacy in this application.}
\label{dm-tab:privacy1}
\end{subtable}

\begin{subtable}{0.48\textwidth}
\begin{tabular}{p{0.45in}<{\centering}p{0.6in}<{\centering}p{0.5in}<{\centering}p{0.6in}<{\centering}p{0.5in}<{\centering}}
\toprule
\multirow{2}{*}{\shortstack{Privacy \\ preference}} &  \multicolumn{2}{c}{MI} & \multicolumn{2}{c}{SIM} \\\cline{2-5}
 & Target Object & Background & Target Object & Background \\\hline
($L_0^t$, $L_0^b$) & 0.00\% & 0.00\% & 0.54 & 0.46 \\
($L_0^t$, $L_1^b$) & 0.02\% & 0.33\% & 0.54 & 0.55 \\
($L_0^t$, $L_2^b$) & 2.48\% & 99.21\% & 0.54 & 0.56 \\
($L_1^t$, $L_0^b$) & 0.46\% & 0.39\% & 0.56 & 0.49 \\
($L_1^t$, $L_1^b$) & 0.42\% & 0.68\% & 0.55 & 0.56 \\
($L_1^t$, $L_2^b$) & 2.97\% & 98.48\% & 0.55 & 0.58 \\
($L_2^t$, $L_2^b$) & 3.45\% & 98.48\%  & 0.64 &  0.58 \\
\bottomrule
\end{tabular}
\caption{Human dataset. Note that the target object in this dataset is the senior person, and the background is a home environment. We assume that the user mainly cares about the target object privacy in this application.}
\label{dm-tab:privacy2}
\end{subtable}
\begin{subtable}{0.48\textwidth}
\begin{tabular}{p{0.45in}<{\centering}p{0.6in}<{\centering}p{0.5in}<{\centering}p{0.6in}<{\centering}p{0.5in}<{\centering}}
\toprule
\multirow{2}{*}{\shortstack{Privacy \\ preference}} &  \multicolumn{2}{c}{MI} & \multicolumn{2}{c}{SIM} \\\cline{2-5}
 & Target Object & Background & Target Object & Background \\\hline
($L_0^t$, $L_0^b$) & 0.00\% & 0.00\% & 0.65 &  0.50 \\
($L_0^t$, $L_1^b$) & 0.00\% & 0.30\%   & 0.65 &  0.53 \\
($L_0^t$, $L_2^b$) & 0.24\% & 99.99\%   & 0.65 &  0.59 \\
($L_1^t$, $L_0^b$) & 0.07\% & 0.03\%  & 0.63 &  0.51 \\
($L_1^t$, $L_1^b$) & 0.04\% & 0.30\%    & 0.66 &  0.59 \\
($L_1^t$, $L_2^b$) & 0.24\% & 99.99\%    & 0.65 &  0.60 \\
($L_2^t$, $L_2^b$) & 0.28\% & 99.99\%   & 0.84 &  0.60 \\
\bottomrule
\end{tabular}
\caption{Bottle dataset. Note that the target object is the pill bottle, and the background is a home environment. We assume that the user mainly cares about the target object privacy in this application.}
\label{dm-tab:privacy3}
\end{subtable}

\begin{subtable}{0.48\textwidth}
\centering
\begin{tabular}{p{0.45in}<{\centering}p{0.6in}<{\centering}p{0.6in}<{\centering}p{0.5in}<{\centering}p{0.5in}<{\centering}}
\toprule
\multirow{2}{*}{Dataset} &  \multicolumn{2}{c}{Target Object} & \multicolumn{2}{c}{Background} \\\cline{2-5}
 & Baseline &\ToolX & Baseline &\ToolX  \\\hline
Husky  & 0.2393 & 0.2695 & 0.2270 & 0.2418 \\
Human & 0.2049  & 0.2429 & 0.2049 & 0.2526 \\
Bottle & 0.2322 & 0.2371 & 0.2088 & 0.2856 \\
\bottomrule
\end{tabular}
\caption{Privacy leakage from text prompt. Note that we compute the SIM metric between text prompts and raw images and use an empty prompt as the baseline.}
\label{dm-tab:privacy4}
\end{subtable}
\caption{Privacy evaluation results. Note that MI measures the mutual information between the sanitized references images shared with the server and the raw reference images. We report the MI w.r.t target object and background separately, and the MI value is normalized by the entropy of the raw image (see Section \ref{subsec:privacy}). SIM measures cosine similarity between the semantic embedding's of the user's private images and the synthetic images generated by the server. Higher values of SIM indicate that more privacy information is being leaked.}
\label{dm-tab:privacy}
\vspace{-0.15in}
\end{table}

\begin{figure*}[htbp]
\centering
\begin{subfigure}{0.24\textwidth}
\includegraphics[width=\linewidth]{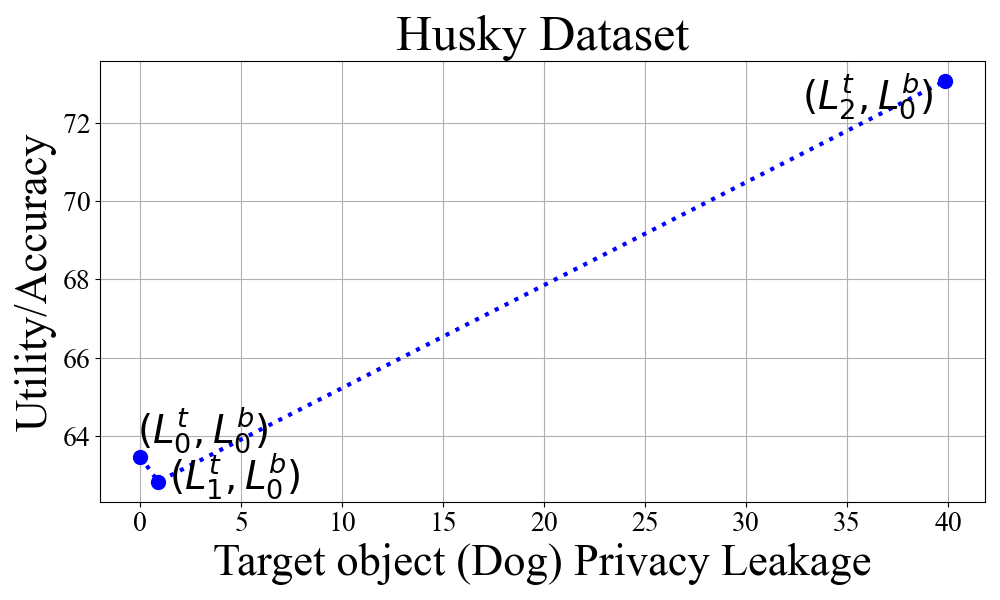}
\caption{MI, Target.}
\label{dm-fig:sub1}
\end{subfigure}
\hfill 
\begin{subfigure}{0.24\textwidth}
\includegraphics[width=\linewidth]{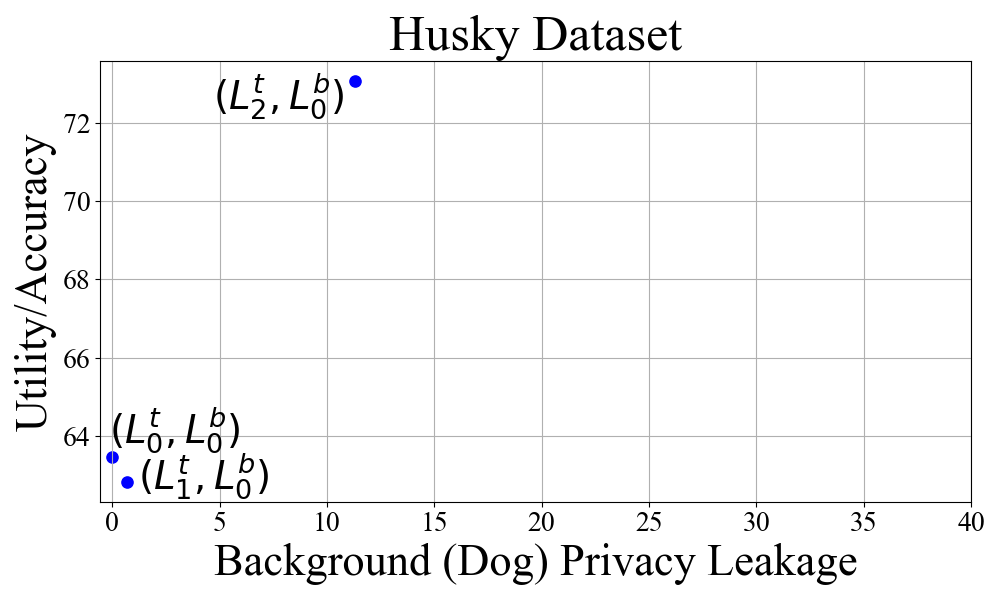}
\caption{MI, Background.}
\label{dm-fig:sub2}
\end{subfigure}
\hfill 
\begin{subfigure}{0.24\textwidth}
\includegraphics[width=\linewidth]{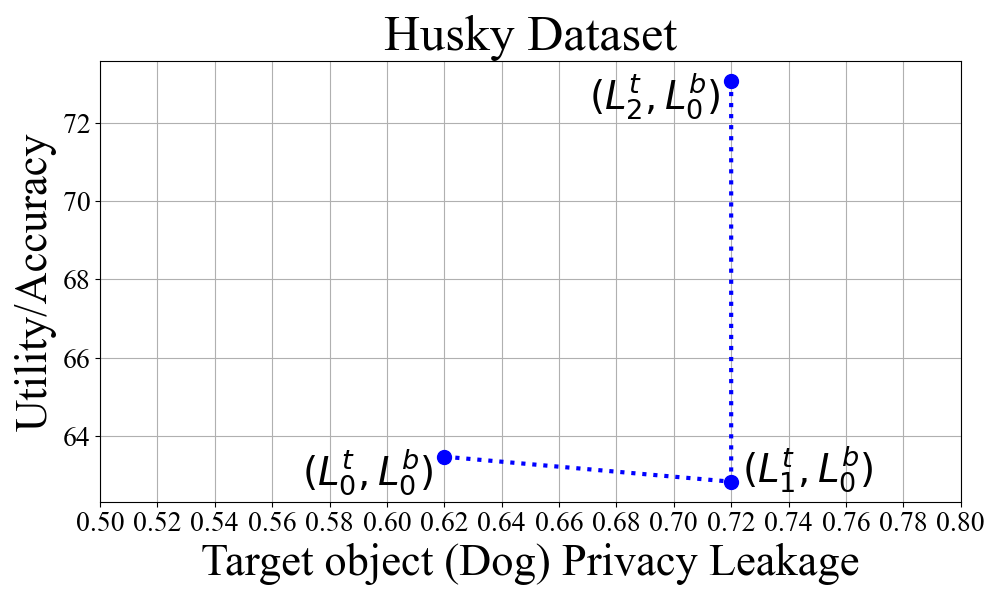}
\caption{SIM, Target.}
\label{dm-fig:sub3}
\end{subfigure}
\hfill 
\begin{subfigure}{0.24\textwidth}
\includegraphics[width=\linewidth]{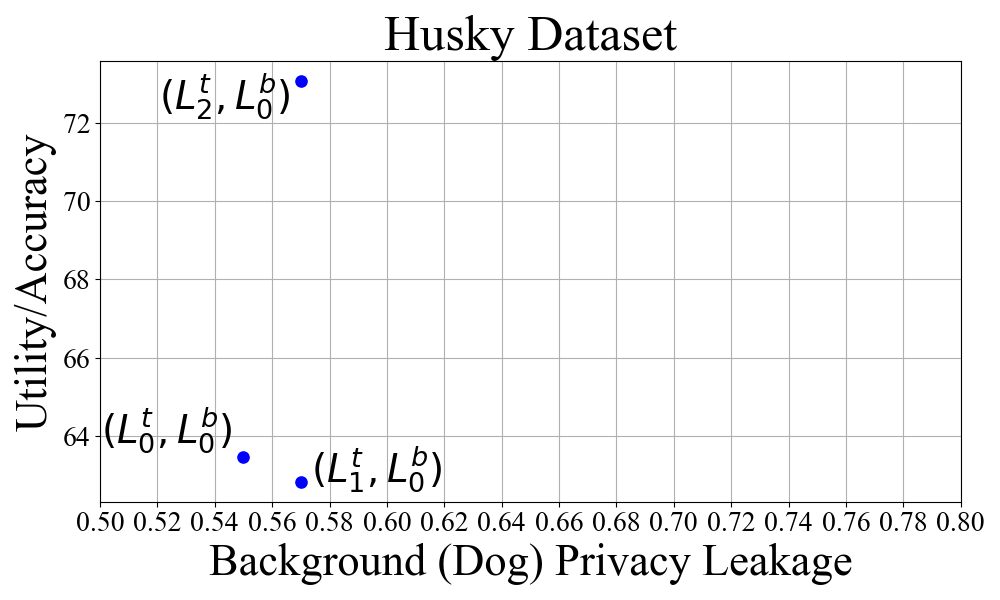}
\caption{SIM, Background. }
\label{dm-fig:sub4}
\end{subfigure}
\begin{subfigure}{0.24\textwidth}
\includegraphics[width=\linewidth]{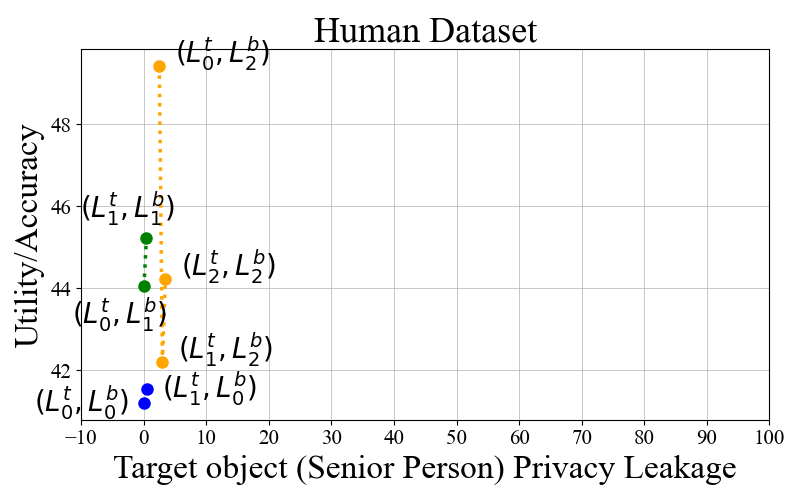}
\caption{MI, Target.}
\label{dm-fig:sub5}
\end{subfigure}
\hfill 
\begin{subfigure}{0.24\textwidth}
\includegraphics[width=\linewidth]{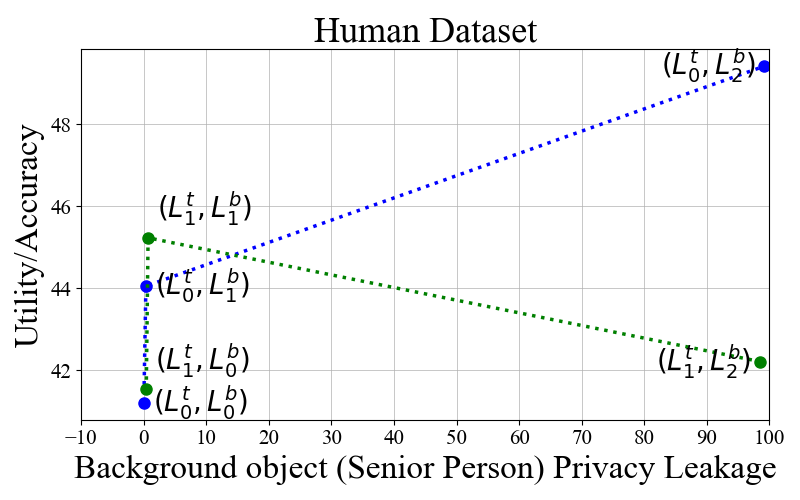}
\caption{MI, Background.}
\label{dm-fig:sub6}
\end{subfigure}
\hfill 
\begin{subfigure}{0.24\textwidth}
\includegraphics[width=\linewidth]{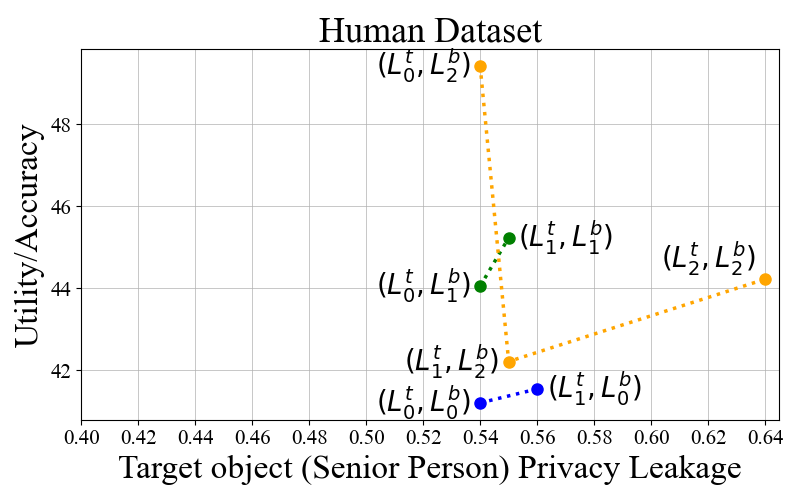}
\caption{SIM, Target.}
\label{dm-fig:sub7}
\end{subfigure}
\hfill 
\begin{subfigure}{0.24\textwidth}
\includegraphics[width=\linewidth]{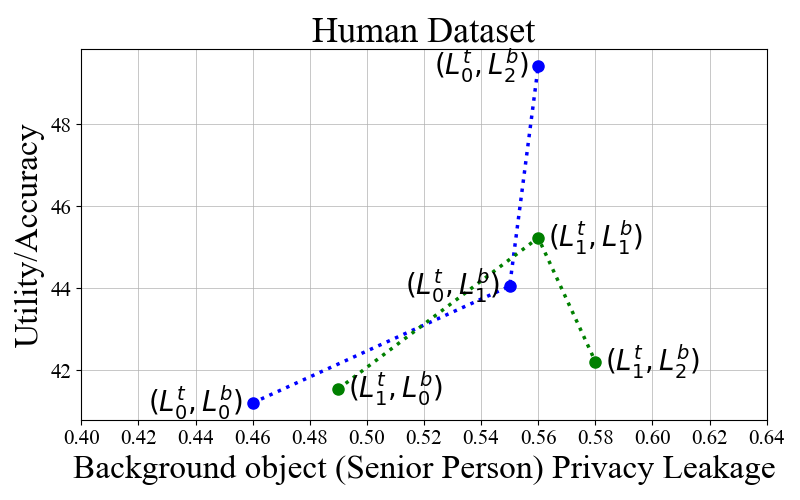}
\caption{SIM, Background. }
\label{dm-fig:sub8}
\end{subfigure}
\begin{subfigure}{0.24\textwidth}
\includegraphics[width=\linewidth]{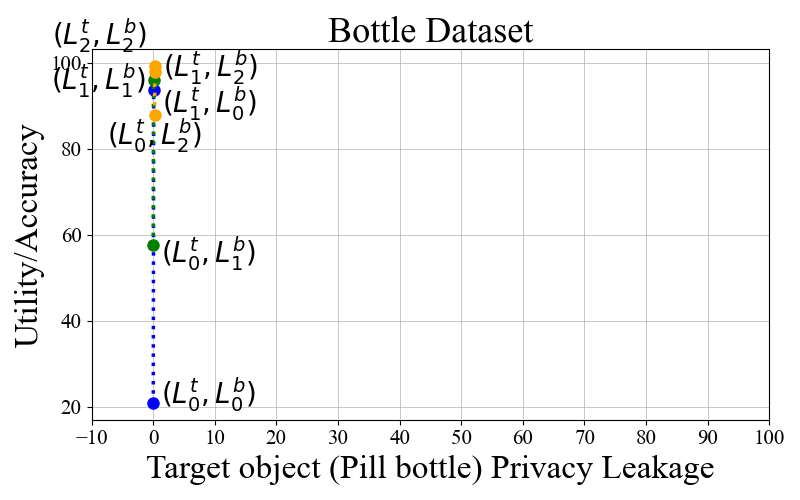}
\caption{MI, Target.}
\label{dm-fig:sub9}
\end{subfigure}
\hfill 
\begin{subfigure}{0.24\textwidth}
\includegraphics[width=\linewidth]{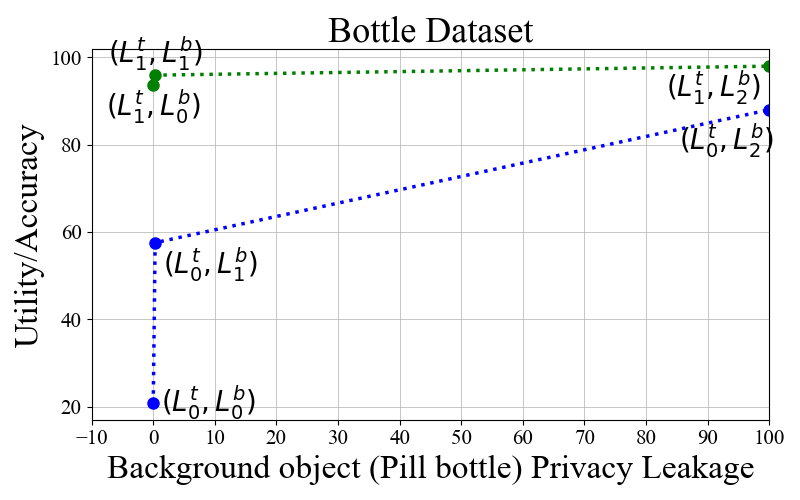}
\caption{MI, Background.}
\label{dm-fig:sub10}
\end{subfigure}
\hfill 
\begin{subfigure}{0.24\textwidth}
\includegraphics[width=\linewidth]{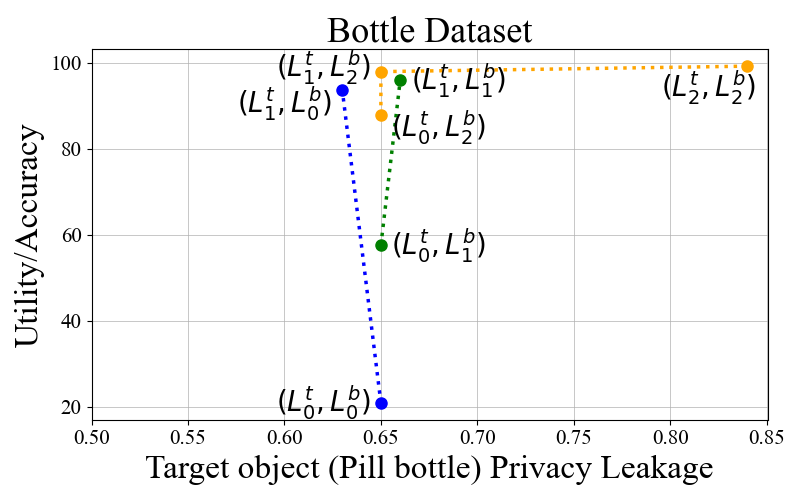}
\caption{SIM, Target.}
\label{dm-fig:sub11}
\end{subfigure}
\hfill 
\begin{subfigure}{0.24\textwidth}
\includegraphics[width=\linewidth]{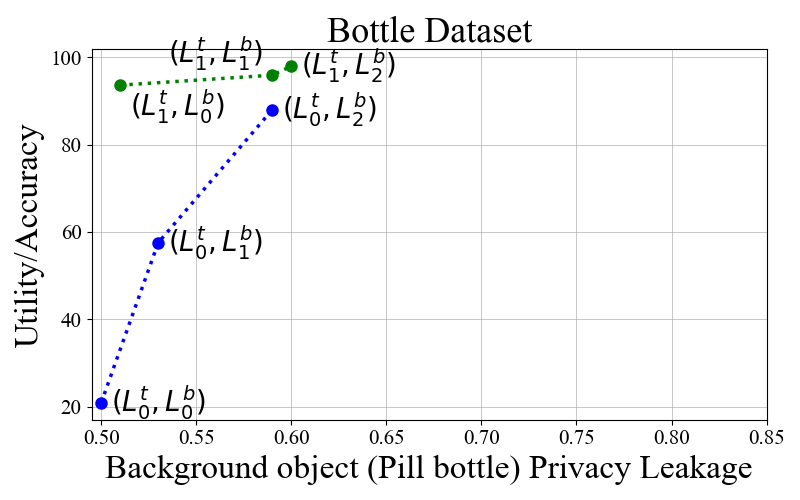}
\caption{SIM, Background. }
\label{dm-fig:sub12}
\end{subfigure}
\caption{Privacy-utility trade-off results. Note that privacy leakage is measured by MI and SIM, and we report leakage w.r.t. Target object and Background separately. 
The model utility represents the performance of the specialized model trained on synthetic data. The top-left part of these figures indicates both higher privacy and higher utility.
Note that lines of a certain color, when present, connecting various points in the graphs illustrate the effect on the privacy-utility trade-off by fixing either the target object or background privacy preference while varying the other. 
}
\label{dm-fig:test}
\end{figure*}

\subsection{Privacy-Utility Trade-offs}
\label{subsec:putrade}
In this subsection, we compare the privacy-utility trade-off performance of customized synthetic data generated with various user privacy preferences in Figure \ref{dm-fig:test}. Note that the privacy leakage is measured by MI and SIM, which represents the mutual information between the sanitized reference images and the raw reference images, and the semantic similarity between generated synthetic images and the user’s private images respectively. The model utility represents the performance of the specialized model trained on synthetic data. The top-left part of these figures indicates both higher privacy and higher utility. \\
\textbf{Husky dataset.} As illustrated in Figures \ref{dm-fig:sub1}-\ref{dm-fig:sub4}, $(L_0^t,L_0^b)$ provides the user with the highest privacy for both the target object and the background (i.e. both MI and SIM privacy leakage are the smallest). This is expected since no reference images or image features are shared with the server under the $(L_0^t,L_0^b)$ privacy preference. In contrast, we observe that the synthetic data generated by $(L_2^t,L_0^b)$ have the lowest privacy but the highest utility, since the raw target object (i.e. husky dog) is shared with the server for customized synthetic data generation. Moreover, as we change the sanitization level of the target object from $L_0$ to $L_2$, the privacy leakage of the target object significantly increases, while the privacy leakage of the background is not significantly changed, as expected. Since the sanitization level of the background is fixed at $L_0$, changing the sanitization level of the target object should not affect the privacy leakage of the background. That said, note that in practice we still observe a slight increase of MI privacy leakage in the background (see Figure \ref{dm-fig:sub2}). This is mainly because the segmentation module cannot perfectly segment the target object from the background, thereby injecting some correlation among the target object and the background.  It is also worth noting that from $(L_0^t,L_0^b)$ to $(L_2^t,L_0^b)$, the user trades the privacy of the target object (husky dog) for better model accuracy without significantly hurting the background privacy. How to choose the privacy and utility trade depends on the user's preference in practice.
\\
\textbf{Human dataset.} In Figures \ref{dm-fig:sub5}-\ref{dm-fig:sub8}, we can see that, as expected, $(L_0^t,L_0^b)$ provides the user with the highest privacy for both the target object (the senior person) and the background (home environment) on human dataset, while having the worst utility. Additionally, when the background sanitization level is fixed, as we change the sanitization level of the target object from  $L0^t$ to $L_1^t$, the privacy leakage of the target object slightly increases while the privacy of the background is not affected. When the sanitization level of the target object is fixed, changing the sanitization level of the background from $L_0^b$ to $L_1^b$ slightly increases the privacy leakage of the background without degrading the privacy of the target object. Moreover, we observe that when the background has sanitization level $L_2^b$ (i.e. $(L_0^t,L_2^b)$ and $(L_1^t,L_2^b)$), both the MI and SIM privacy leakage of the background is significantly higher, since the raw background images are shared with the server. However, this also brings higher utility gain for the user. In practice, by changing the sanitization level of the background from $L_0^b$ to $L_2^b$, the user can trade between the privacy of the background and the utility of the model, without hurting the privacy of the target object (senior person).
\\
\textbf{Bottle dataset.} As shown in Figures \ref{dm-fig:sub9}-\ref{dm-fig:sub12}, $(L_0^t,L_0^b)$ provides the user with the highest privacy for both the target object (the bottle) and the background (bedroom environment), while the utility is the lowest. Changing the sanitization level of the background from $L_0^b$ to $L_2^b$ increases the privacy leakage of the background, while not affecting the privacy of the target object. Moreover, this brings significant utility gain. In practice, if the user mainly cares about the privacy of the pill bottle, then the user can choose to trade the privacy of the background for higher model accuracy. 

\subsection{Key Insights from Privacy-Utility Analysis}
\textbf{Insight 1: $L_0\rightarrow L_2$ may increase utility with limited privacy decrease.}
We observe that, as expected, increasing the sanitization level from $L_0$ to $L_2$ leads to less privacy and more utility. Specifically, compared with $L_0$, using $L_1$ on the target object or background will slightly increase the privacy leakage while using $L_2$ on the target object or background will significantly increase the privacy leakage. However, varying the sanitization level of the target object from $L_0$ to $L_2$ will not significantly change the privacy leakage of the background, and vice versa. This allows users to effectively balance between privacy and utility, in use cases where only some parts of the image data are sensitive, see, for example, the $(L_2^t,L_0^b)$ point in Figures \ref{dm-fig:sub2} and \ref{dm-fig:sub4} in the husky dog use case, and the $(L_0^t,L_2^b)$ point in Figures \ref{dm-fig:sub5} and \ref{dm-fig:sub7} in the senior person use case.
\\
\noindent
\textbf{Insight 2: Whether $L_1$ scheme offers utility gain or not depends on the application.}
When $L_1$ is used, which means additional features are shared with the server, whether the customized synthetic data can improve the model accuracy depends on whether the DM can be properly fine-tuned on the shared features. 
Consider the husky dataset. The shared edge images of the target objects (i.e. husky dogs) are not associated with labels, hence the DM does not have information about the correlation between different canny edge images and the status of the dog. For example, the canny edge of a sleeping dog will be different from the canny edge of a dog which is playing. Without knowing the label of each canny edge shared with the server, the DM may use the canny edge of a sleeping dog to generate a dog which is playing. In this case, the DM cannot be fine-tuned properly and hence sharing the canny edge does not benefit the training of the specialized ML model. 
Similarly, for the human dataset, the shared pose images are also not associated with labels. In this case, the DM has limited information about which specific activities the pose represents, and thus the specialized model accuracy is not significantly improved using the customized synthetic data. 
In contrast, for the bottle dataset, the label is the location of the pill bottle, which can be obtained during the synthetic data generation process (see Section \ref{dm-sec:setup} for details). In this case, the canny edge images of pill bottles can be leveraged properly to fine-tune the DM and thus enhance the accuracy of the specialized ML model, see how the the $(L_1^t,L_i^b)$ points are consistently superior to the $(L_0^t,L_i^b)$ points, $i \in \{0,1,2\}$ in the pill bottle detection use case in Figures \ref{dm-fig:sub9}-\ref{dm-fig:sub12}.
\\
\noindent
\textbf{Insight 3: Users can trade between privacy and utility differently for different applications. }
The proposed \ToolX~effectively allows users to choose their level of privacy based on utility gains in different situations, as motivated by Table \ref{tab:motivation} in Section \ref{subsec:problem}. If the user cares more about utility in an application, then they can choose to trade privacy on less sensitive objects in order to gain more utility. In contrast, if the privacy of certain objects is more important to users, the utility provided by these objects needs to be traded for privacy. Figures \ref{dm-fig:sub1}-\ref{dm-fig:sub12} depict a number of situations where such privacy-utility trades take place, and also highlight use cases where privacy can be achieved without a sizable loss in utility.  
\subsection{Sanitization via Noise Addition}
\label{subsec:obfuscation}
Noise addition is a well established method to protect privacy (e.g. \cite{zhang2022privacy,dwork2014algorithmic}). In this subsection and in contrast to the previous sanitization methods, we explore adding random noise on the user's reference images. Specifically, we consider the Husky dataset and add Gaussian noise with varying variance $\sigma^2$ to the target object (i.e. husky dog) in the user's reference images. We use the noisy reference images to fine-tune the DM model to generate customized husky dog images, train the specialized model, and then test the model accuracy versus utility. To explore how different amount of noise affects the privacy and utility, we add Gaussian noise with three different levels of variance: low ($\sigma=5$), middle ($\sigma=10$), and high ($\sigma=50$). Lower variance indicates less obfuscation noise and hence less privacy (larger $\epsilon$ value in the DP bound \cite{dwork2014algorithmic}), while higher variance indicates more obfuscation noise and hence more privacy (smaller $\epsilon$ value).

\begin{table}[ht]
\centering
\begin{tabular}{p{0.6in}<{\centering}p{0.85in}<{\centering}p{0.85in}<{\centering}p{0.5in}<{\centering}}\\\toprule
\multirow{2}{*}{Variance} & \multicolumn{2}{c}{Privacy leakage (MI)} & Utility \\\cline{2-3}
& MI & SIM &  Accuracy \\\hline
Low & (15.10\%, 5.07\%) & (0.71, 0.58) & 44.23\%\\
Mid & (12.84\%, 4.77\%) & (0.69, 0.58) & 42.31\% \\
High & (9.45\%, 4.55\%) & (0.65, 0.58) & 42.31\% \\
\hline
N/A ($L_0$) & (0.00\%, 0.00\%) & (0.62, 0.55) & 63.46\%\\
\bottomrule
\end{tabular}
\caption{Privacy-utility trade-off results of adding obfuscation noise on husky dataset. Note that ($x,y$) in the MI/SIM privacy column means MI/SIM privacy leakage of target object and background respectively, and higher values indicate more privacy leakage and hence less privacy. The last row is the results of \ToolX~when the privacy preference is set as $(L_0^t,L_0^b)$. We use the accuracy of specialized ML models as the utility metric.}
\label{dm-table:noise}
\vspace{-0.1in}
\end{table}
As reported in Table \ref{dm-table:noise}, when the variance of Gaussian noise increases from low to high value, the privacy leakage with respect to both MI and SIM will decrease, while the utility also decreases, as expected. The fourth row in the table repeats, for comparison purposes, the MI/SIM privacy leakage and model accuracy when the $L_0$ sanitization scheme is applied on the target object. (Recall that under $L_0$ we only share the deterministic text description and the MI leakage in this case is zero.) Notably, sharing a reference image where we add noise on the target object causes the utility to drop significantly as compared to $L_0$, while there is no privacy benefit: the privacy leakage increases since, in addition to sharing the text, we also share a noisy image. From the model accuracy results it is evident that DM fine-tuning based on a noisy image is counterproductive.

In terms of DP guarantees, even for the largest value of $\sigma$ considered, which will yield the smallest $\epsilon$, the corresponding $\epsilon$ value is very large (larger than one thousand due to the high associated sensitivity of the dataset \cite{dwork2014algorithmic}) which represents a completely meaningless value in terms of the DP bound. Last, while there are more efficient methods to apply noise in images for DP  purposes, e.g. adding the noise in a latent space \cite{kim2019latent}, intuitively,  DP's requirement to guarantee a privacy bound under any distribution will make it very hard, if not impossible, to get both good privacy and usable utility in our setting when an image segment is both sensitive and of sizable utility.

Therefore, we conclude that adding obfuscation noise to reference images will significantly decrease the utility of the generated customized synthetic data and make them unusable in practice, without offering any practical privacy gains over the proposed sanitization schemes.

\subsection{Generalization to multiple objects}
In this subsection, we conduct a case study by extending the non-popular object detection task when there are three image segments: pill bottle, photo frame, and bedroom background. We consider the pill bottle and photo frame as sensitive objects. We report the detection accuracy of the pill bottle with mAP50 as the utility evaluation metric while applying sanitization levels $L_0$ and $L_1$ for both sensitive objects. 

\begin{table}[ht]
\centering
\small
\begin{tabular}{p{1.2in}<{\centering}p{0.8in}<{\centering}p{0.8in}<{\centering}}
\hline
\multirow{2}{*}{Privacy preference} &  \multicolumn{2}{c}{Accuracy (mAP50)}\\\cline{2-3}
 & Synthetic data & Real-word data \\\hline
($L_0^{t1}$, $L_0^{t2}$, $L_0^b$) & 93.02\% & 44.68\% \\
($L_0^{t1}$, $L_0^{t2}$, $L_1^b$) & 92.82\% & 71.95\% \\
($L_0^{t1}$, $L_0^{t2}$, $L_2^b$) & 94.04\% & 72.18\% \\

($L_0^{t1}$, $L_1^{t2}$, $L_0^b$) & 93.74\% & 42.84\% \\
($L_0^{t1}$, $L_1^{t2}$, $L_1^b$) & 94.26\% & 53.06\% \\
($L_0^{t1}$, $L_1^{t2}$, $L_2^b$) & 94.49\% & 77.78\% \\

($L_1^{t1}$, $L_0^{t2}$, $L_0^b$) & 83.69\% & 93.56\% \\
($L_1^{t1}$, $L_0^{t2}$, $L_1^b$) & 84.75\% & 88.93\% \\
($L_1^{t1}$, $L_0^{t2}$, $L_2^b$) & 83.80\% & 94.41\% \\

($L_1^{t1}$, $L_1^{t2}$, $L_0^b$) & 78.73\% & 92.07\% \\
($L_1^{t1}$, $L_1^{t2}$, $L_1^b$) & 87.82\% & 93.78\% \\
($L_1^{t1}$, $L_1^{t2}$, $L_2^b$) & 71.15\% & 94.61\% \\
($L_2^{t1}$, $L_2^{t2}$, $L_2^b$) & 83.65\% & 99.47\% \\
\hline
\end{tabular}
\caption{Utility evaluation results for two objects dataset which contains images of a medicine pill bottle and personal photo frame randomly located in a bedroom. Note that we use mAP50 (mean average precision calculated at an intersection over union (IoU) threshold of 0.50) as the utility metric for the two objects dataset.
}
\label{dm-tab-2-obj:utility-6}
\vspace{-0.1in}
\end{table}

As reported in Table \ref{dm-tab-2-obj:utility-6}, when varying the sanitization level of the background from $L_0^b$ to $L_2^b$ while applying sanitization level $L_0$ to both the pill bottle and photo frame, the utility increases while maintaining privacy for the sensitive target objects. When varying the sanitization level of the second sensitive object (photo frame) for varying sanitization levels of the background, we can see that the utility remains consistent indicating that the second target object doesn't affect the accuracy of the pill bottle detection. 
In contrast, when the sanitization level of the pill bottle is increased from $L_0^{t_1}$ to $L_1^{t_1}$ irrespective of the sanitization levels for the second target object and background, the utility improves significantly.

The privacy leakage of the bill bottle and background are not affected by the addition of the photo frame and remain the same as reported in Table \ref{dm-tab:privacy3}.
We compute the privacy leakage of the photo frame when varying its sanitization level using the MI metric.  We find the privacy leakage to be small, specifically 
0.00\% for the $L_0$ and 1.43\% for the $L_1$ scheme. This is consistent with the privacy leakage results for the pill bottle as reported in Table \ref{dm-tab:privacy3}.

These findings suggest that for users primarily interested in maintaining privacy for the target objects, there is an opportunity to balance privacy and utility by adjusting the sanitization levels for the background. By allowing slightly higher privacy leakage for non-sensitive components (background), it is possible to achieve higher model accuracy for detecting the target objects. This emphasizes that a strategic trade-off between sanitization levels across different image segments can lead to better detection performance while still maintaining a reasonable level of privacy for sensitive objects.

\begin{table}[b]
\centering
\begin{tabular}{p{1.2in}<{\centering}p{1.5in}<{\centering}}
\hline
Dataset &  Accuracy/mAP50\\\hline
Husky & 0.00\% \\
Human & 1.60\% \\
Bottle & 17.30\% \\
\hline
\end{tabular}
\caption{SEEM's performance on three datasets. Note that we report accuracy of SEEM on Husky and Human dataset, and mAP50 of SEEM on Bottle dataset. 
}
\label{dm-tab:utility-seem}
\vspace{-0.2in}
\end{table}

\subsection{Using generic segmentation/detection models} 
\label{sub-section-5.7:comparison-with-seem}
We investigate how a large foundation model which specializes in object segmentation and detection may perform in the three use cases we consider, without considering the practical challenges from using such models as described in Section \ref{genericmodel}, i.e. fixed output format, weak generalization to personalized data, and prohibitively large size for end devices. We select SEEM \cite{zou2023seem} as it is a state of the art model for segmentation/detection.

In the Husky dataset where dogs are categorized as eating, sitting, sleeping, or playing, and the Human dataset were human activities are classified as eating, drinking, walking and reading, we adapted the SEEM model, which is primarily designed for object segmentation/detection, to evaluate its capability in detecting objects based on their specific states. Specifically, we modified the text prompts given to the SEEM model to include information about the state of the object.
SEEM achieved detection scores of 97\% and 98\%, respectively. Yet, it exhibits limitations when classifying objects based on specific states, e.g. dog sitting versus sleeping, resulting in low classification accuracy as seen in Table \ref{dm-tab:utility-seem}. This is consistent with our intuition that such models may not perform well unless they are fine-tuned to learn personalized tasks, which is very expensive to do in practice.

In the Bottle dataset focusing on pill bottles in a bedroom, SEEM's detection score was 17.30\%, which was notably low compared to a dataset of beer bottles in a bedroom, where it achieved 99\% accuracy. This disparity underscores SEEM's challenges in detecting uncommon objects. 

In contrast to SEEM, \ToolX~ addresses these limitations by fine-tuning smaller, task-specific models that are optimized for performance on resource-constrained devices. This adaptability makes \ToolX~ a valuable tool for personalized ML model training and deployment, especially in environments with limited computational resources.

\vspace{-0.05in}
\section{Related Work}
\noindent\textbf{Diffusion Models.} The diffusion model (DM) was first introduced by Sohl et al. \cite{sohl2015deep}. It involves a forward diffusion process that incrementally adds noise to data, and a reverse diffusion process that reconstructs the original data from noise. Later, Jonathan et al. \cite{ho2020denoising} demonstrated that DMs can efficiently generate high-quality synthetic images, surpassing previous methods such as Variational Autoencoders (VAEs) \cite{kingma2013auto} and Generative Adversarial Networks (GANs) \cite{goodfellow2014generative}. To generate high-resolution synthetic images, Latent DMs (also known as Stable Diffusion Models) were proposed by Rombach et al. \cite{rombach2022high}, conducting diffusion and denoising processes in latent space. Additionally, various conditioning mechanisms introduced in \cite{rombach2022high} have transformed DMs into flexible conditional image generators, supporting applications like text-to-image and super-resolution image generation. Recent advancements have incorporated specific conditioning during the denoising phase to align synthetic images more closely with reference images in terms of edges, depth, and structure \cite{zhao2023uni, zhang2023adding, huang2023fine}, fostering more controlled and realistic image generation.

\noindent\textbf{Synthetic Data Generation.} Extensive research has demonstrated that combining synthetic data with real data can enhance the performance of machine learning models across critical vision and control applications, such as image classification, semantic segmentation, face recognition, and autonomous vehicle control \cite{sankaranarayanan2018learning, nikolenko2021synthetic, bousmalis2018using, osokin2017gans, muller2018driving, wood2021fake, azizi2023synthetic, Sariyildiz_2023_CVPR, zhou2023training, he2022synthetic,yang2023diffusion}. These studies have employed a range of generative models, from GANs \cite{goodfellow2014generative} to Stable Diffusion \cite{rombach2022high}, to create synthetic datasets for model training. While previous efforts have primarily focused on using synthetic data to complement real-world training data for improving model performance, our work investigates scenarios where users need to train specialized ML models on specific tasks involving private data distributions, and where labeled real-world data is unavailable. Therefore, customized synthetic data needs to be generated for training specialized ML models. To achieve the goal of tailoring the distribution of generated synthetic data, recent works have proposed various methods, including fine-tuning the models on a set of reference images \cite{ruiz2023dreambooth} or incorporating conditional reference image features to the image generation process \cite{zhang2023adding}. Different from these works, we further study the privacy leakage problem when customizing the synthetic data generation process, and propose a novel framework that allows user to balance the privacy and utility of customized synthetic data, based on their privacy preferences.

\noindent\textbf{Data Obfuscation}
Prior works have explored various data obfuscation methods, which add noise into user data to protect their private information. For example, \cite{infotheor_util,info_theo,matt} explored the addition of information-theoretical-based noise into real-world user data to provide on-average privacy guarantees. The authors in \cite{huang2018generative,huang2017context,raval2019olympus,zhang2022privacy} proposed to inject generative adversarial privacy (GAP) noise into user data, which may exhibit better on-average privacy-utility trade-offs on many real-world applications. \cite{localdp, dp_lopub, fan2018image} designed different DP noise addition mechanisms, which add DP noise into user data to achieve worst-case privacy guarantees.  In \cite{abadi2016learning,xue2021dp,kim2019latent}, researchers have explored the usage of Encoder-Decoder neural networks to map the image data into a latent vector and then add noise into the latent vector, which achieves better privacy-utility trade-offs compared with adding noise to image data directly.  Note that the data obfuscation methods proposed in these works are orthogonal to our proposed \ToolX, since they can be employed as additional sanitization schemes on top of the image sanitizer in \ToolX. As an example, we explore how Gaussian DP mechanism can be used as an additional sanitization scheme in \ToolX, though it may lead to low accuracy as we demonstrated in Section \ref{subsec:obfuscation}. 




\noindent\textbf{Privacy-preserving Machine Learning.}
Prior works have investigated how to train ML models at a server without collecting users' private local data, using federated learning (FL) \cite{mcmahan2017communication, kairouz2021advances}. In FL, instead of sharing private data to the server, users train the model locally on their private data and then send the local model updates to the server. Recent works (e.g., \cite{qi2024tomtit}) have proposed efficient mechanisms to address the challenge of applying FL to large-scale model training. However, the model updates shared with the server in FL can also leak privacy \cite{yin2021gradients, geiping2020inverting}. To mitigate this privacy leakage, a number of privacy-enhancing mechanisms have been proposed for FL, which include homomorphic encryption \cite{aono2017privacy}, secure aggregation \cite{secagg_so2021securing,so2021turbo} and adding DP noise into model updates \cite{abadi2016deep,truex2020ldp,el2022differential}. Different from the above works in FL, \ToolX~proposes to privately train specialized models under scenarios where there may be a lack of labeled data required for training, and the user may have a unique request for a personalized task and hence other users’ data may not be leveraged effectively to collaboratively train the corresponding model. Last, note that federated, and, more general, distributed learning, may be applied on top of \ToolX~if there are other users' data that can be collaboratively used for the same training task. 

%
In summary, the key novelty of our work is the design of a general framework which offers users fine-grained, object-level privacy control over the data/images shared with the server. With \ToolX, users can set the sensitivity of image objects/regions differently based on their privacy preferences and flexibly trade the privacy of less sensitive objects for better application-specific utility.

\vspace{-0.05in}
\section{Discussion, Limitations and Future Work}
\label{sec:discussion}

\noindent\textbf{System cost and overhead.} We clarify that the majority of the computational and storage cost from running \ToolX~is on the server side, since the server needs to fine-tune the DMs, generate synthetic training data, and then train the specialized ML models. The user-device only needs to run a light-weight object detection and segmentation model on a few reference images, which has a negligible runtime cost and overhead on local devices \cite{Yolov8Ultralytics}.

\noindent\textbf{Generalization to multi-object scenarios.} Note that \ToolX~ can be generalized to multi-object scenarios, where the user can have different privacy preferences for multiple objects/segments in the image. In this case, the image sanitizer in \ToolX~ will run sanitization on all objects in parallel and then share them with the server. The server will customize the DM to generate each one of these objects and then merge them into the synthetic images.

\noindent\textbf{Other image sanitizers.} \ToolX~ offers a general framework for users to trade between the privacy and utility during private training of specialized models. Other image sanitizers with different sanitization schemes can be easily integrated into \ToolX. Future work may explore how to design more principled image sanitizers, e.g. using Reinforcement Learning and other techniques, to optimize the privacy and utility trade-offs in \ToolX~given appropriate objective functions.

\noindent\textbf{Effects of user labels.} In our current threat model, we assume that the user does not provide any labeled reference images. However, in practice, the user may be willing to label a few reference images and share them with the server in order to obtain a more accurate specialized model.
(Note that the labeling process may degrade the user experience and requires the design of a labeling system.) 
Future work can explore to what extend a few labeled reference images may boost the utility of specialized model training.

\noindent\textbf{Synthetic video data generation.} Our current experiments are limited to using a text-to-image diffusion model for image data generation, due to the lack of open-sourced high-quality video diffusion models and the high runtime cost for video data generation. However, for certain computer vision tasks like human activity monitoring, in order for the ML models to achieve good accuracy, they need to take video data as input \cite{dai2022toyota}. Future work could explore how to leverage video diffusion models for customized synthetic video data generation.

\noindent\textbf{Advanced and future diffusion models.} Our experiments use the Stable-Diffusion-v1.5 model for customized synthetic data generation. In practice, with more computation resources, the server could use larger diffusion models (e.g. Stable-Diffusion XL \cite{sdxl2023}) to generate synthetic data with better quality. Moreover, with the rapid development of generative AI, more advanced and efficient image and video generation models (e.g. GPT-4o \cite{openai2023gpt4}) will appear and hence the customized synthetic data generation performance of \ToolX~ will be further enhanced.

\noindent\textbf{Advanced object detection and segmentation models.} The current design of \ToolX~ is limited to using a pre-trained off-the-shelf object detection and segmentation model which may not always detect and segment objects correctly.
We can improve the robustness of object detection and segmentation on user's private data by leveraging \ToolX~ to generate customized synthetic training data and fine-tune the object detection and segmentation module. Moreover, we expect future object detection and segmentation models will have high accuracy without the need for fine-tuning. 

\vspace{-0.05in}
\section{Conclusion}
In this work, we propose \ToolX, a novel system to generate customized synthetic image data for specialized ML model training, where the user only needs to share a few sanitized reference images. Moreover, the proposed system provides users with fine-grained  and object-level privacy control of sanitized reference images, allowing them to balance privacy and utility according to their preferences. Our experiments across three distinct model training tasks demonstrate that \ToolX~achieves good model accuracy without compromising the privacy of sensitive user information.

\begin{acks}
This work is partially supported by NSF (SaTC-1956435, CNS-1901488) and a USC-Amazon Research Center award.
\end{acks}

\bibliographystyle{ACM-Reference-Format}
\bibliography{ref, related_works}

\appendix
\section{Visualization Results}
\begin{figure*}[h]
\centering
\begin{subfigure}{0.22\textwidth}
\includegraphics[width=\linewidth]{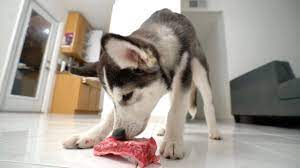}
\caption{Raw image}
\label{dm-fig:husky-ex1}
\end{subfigure}
\hfill 
\begin{subfigure}{0.22\textwidth}
\includegraphics[width=\linewidth]{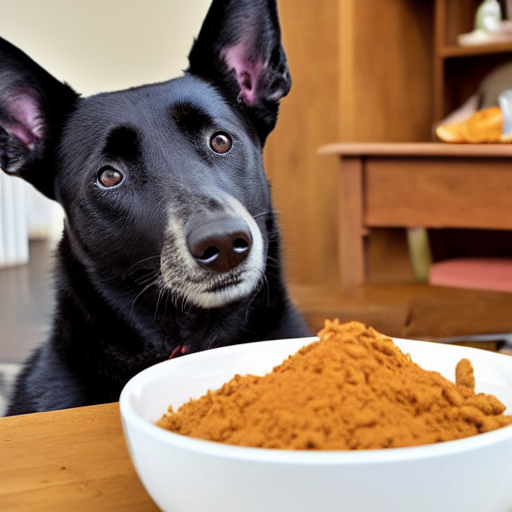}
\caption{($L_0^t$, $L_0^b$)}
\label{dm-fig:husky-ex2}
\end{subfigure}
\hfill 
\begin{subfigure}{0.22\textwidth}
\includegraphics[width=\linewidth]{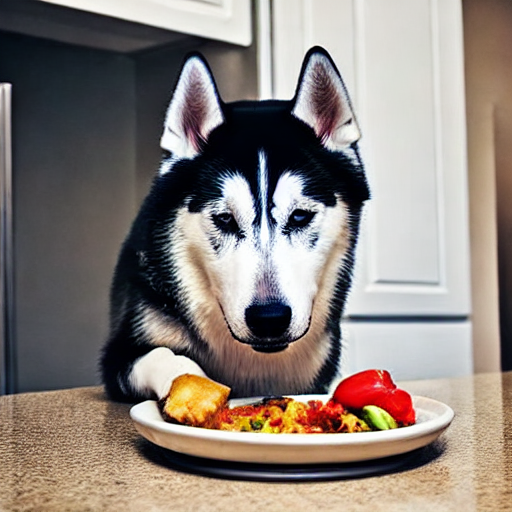}
\caption{($L_1^t$, $L_0^b$)}
\label{dm-fig:husky-ex3}
\end{subfigure}
\hfill 
\begin{subfigure}{0.22\textwidth}
\includegraphics[width=\linewidth]{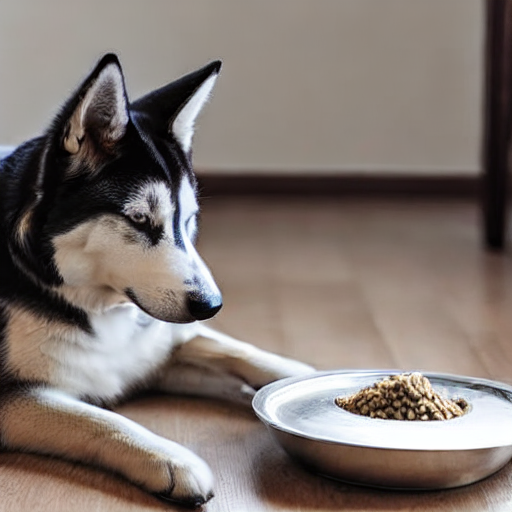}
\caption{($L_2^t$, $L_0^b$)}
\label{dm-fig:husky-ex4}
\end{subfigure}
\caption{Visualization results of Husky dataset.}
\label{dm-fig:husky-examples}
\end{figure*}

\begin{figure*}[!h]
\centering
\begin{subfigure}{0.3\textwidth}
\includegraphics[width=\linewidth]{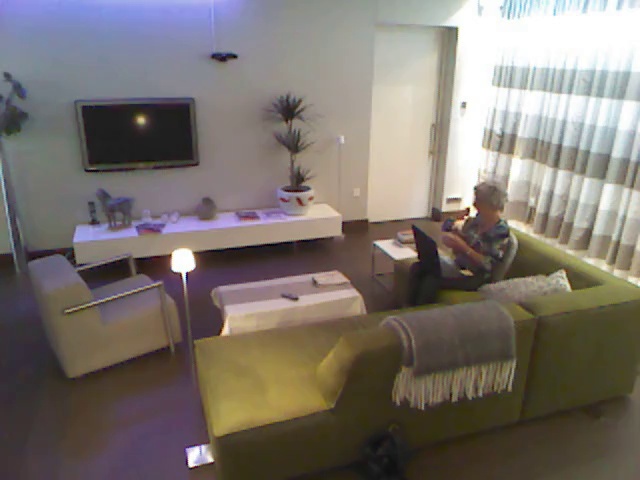}
\caption{Raw image.}
\label{dm-fig:human-ex1}
\end{subfigure}
\hfill 
\begin{subfigure}{0.3\textwidth}
\includegraphics[width=\linewidth]{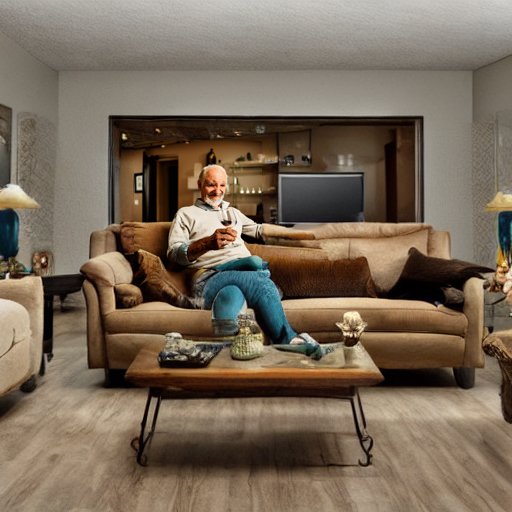}
\caption{($L_0^t$, $L_0^b$)}
\label{dm-fig:human-ex2}
\end{subfigure}
\hfill 
\begin{subfigure}{0.3\textwidth}
\includegraphics[width=\linewidth]{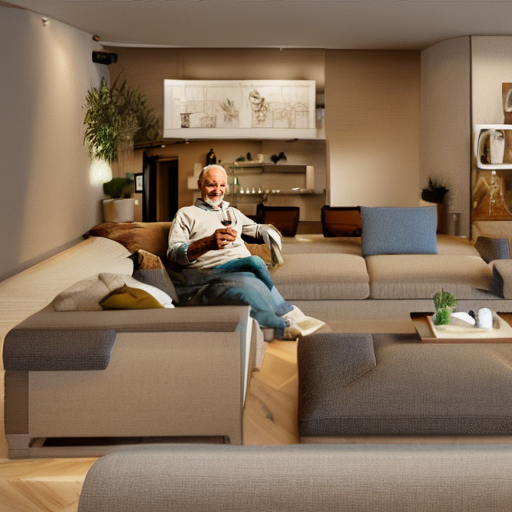}
\caption{($L_0^t$, $L_1^b$)}
\label{dm-fig:human-ex3}
\end{subfigure}
\hfill 
\begin{subfigure}{0.3\textwidth}
\includegraphics[width=\linewidth]{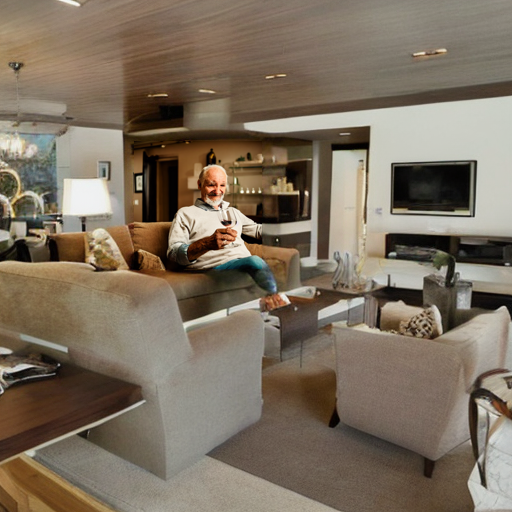}
\caption{($L_0^t$, $L_2^b$)}
\label{dm-fig:human-ex4}
\end{subfigure}
\hfill 
\begin{subfigure}{0.3\textwidth}
\includegraphics[width=\linewidth]{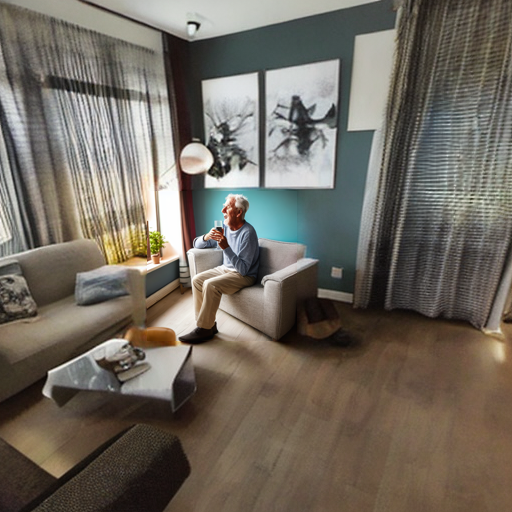}
\caption{($L_1^t$, $L_1^b$)}
\label{dm-fig:human-ex5}
\end{subfigure}
\hfill 
\begin{subfigure}{0.3\textwidth}
\includegraphics[width=\linewidth]{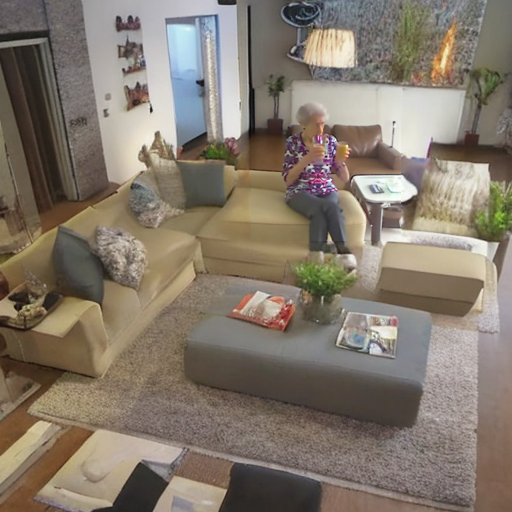}
\caption{($L_2^t$, $L_2^b$)}
\label{dm-fig:human-ex6}
\end{subfigure}
\caption{Visualization results of Human dataset.
}
\label{dm-fig:human-examples}
\end{figure*}

\begin{figure*}[!h]
\centering
\begin{subfigure}{0.3\textwidth}
\includegraphics[width=\linewidth]{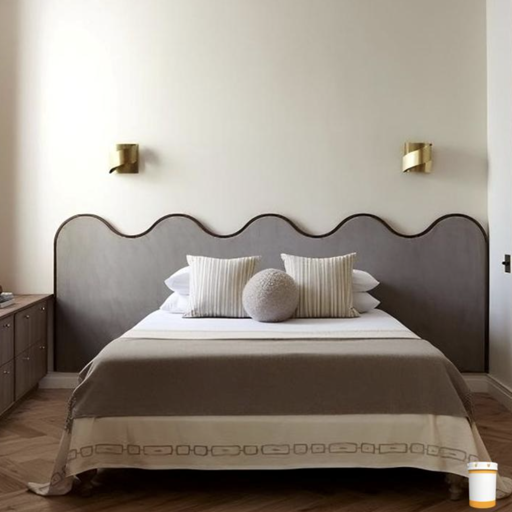}
\caption{Raw image.}
\label{dm-fig:bottle-ex1}
\end{subfigure}
\hfill 
\begin{subfigure}{0.3\textwidth}
\includegraphics[width=\linewidth]{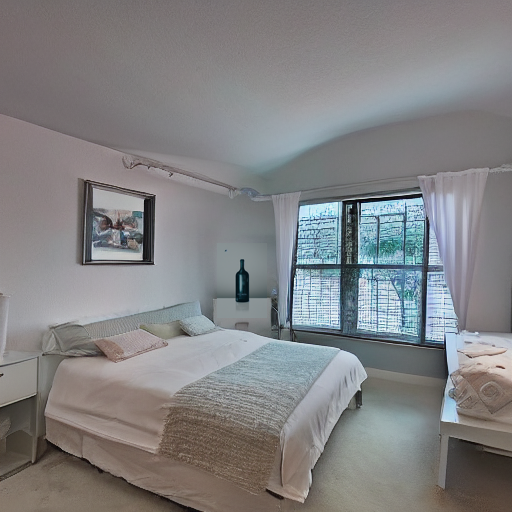}
\caption{($L_0^t$, $L_0^b$)}
\label{dm-fig:bottle-ex2}
\end{subfigure}
\hfill 
\begin{subfigure}{0.3\textwidth}
\includegraphics[width=\linewidth]{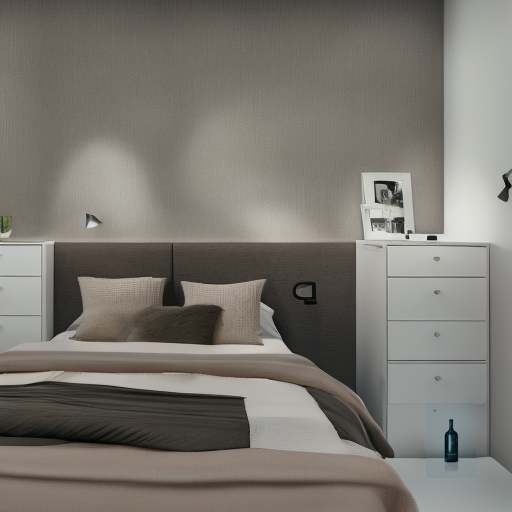}
\caption{($L_0^t$, $L_1^b$)}
\label{dm-fig:bottle-ex3}
\end{subfigure}
\hfill 
\begin{subfigure}{0.3\textwidth}
\includegraphics[width=\linewidth]{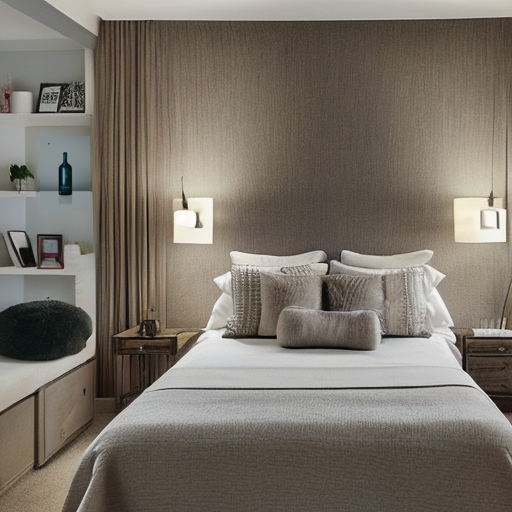}
\caption{($L_0^t$, $L_2^b$)}
\label{dm-fig:bottle-ex4}
\end{subfigure}
\hfill 
\begin{subfigure}{0.3\textwidth}
\includegraphics[width=\linewidth]{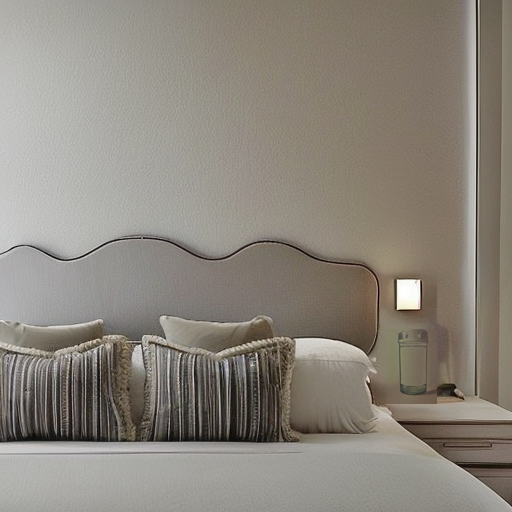}
\caption{($L_1^t$, $L_2^b$)}
\label{dm-fig:bottle-ex5}
\end{subfigure}
\hfill 
\begin{subfigure}{0.3\textwidth}
\includegraphics[width=\linewidth]{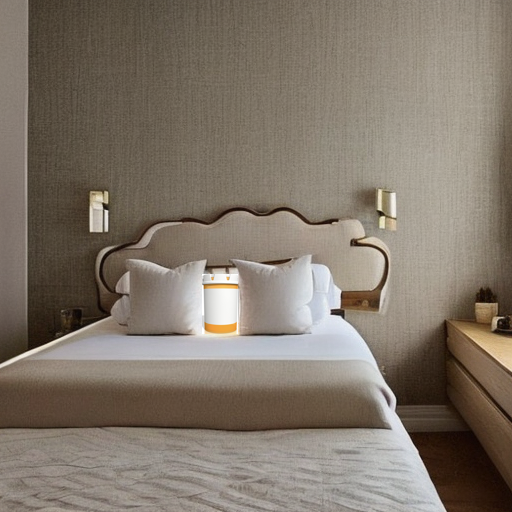}
\caption{($L_2^t$, $L_2^b$)}
\label{dm-fig:bottle-ex6}
\end{subfigure}
\caption{Visualization results of Bottle dataset.
}
\label{dm-fig:bottle-examples}
\end{figure*}

We provide some examples of the generated synthetic images under different sanitization schemes in Figure \ref{dm-fig:husky-examples}-\ref{dm-fig:bottle-examples}.

\section{Model Training Configurations}
\subsection{DM Fine-tuning}
We select Stable-Diffusion-v1.5 \cite{rombach2022high} as our DM, and we fine-tune it using DreamBooth algorithm \cite{ruiz2023dreambooth} to generate both customized target object and the background during our experiments. We use the diffusers library from Huggingface \cite{HuggingfaceDiffusers} to fine-tune DM. Specifically, we set learning rate as 2e-6, the specical token as \textbf{xyz->style}, the prior loss weight as 0.01, and the gradient accumulation step as 2 in DreamBooth algorithm. We fine-tune the model for 800 steps.
\subsection{Specialized ML Model Training}
\noindent\textbf{MobileNet.} For the pet status and human activity monitoring tasks, we use MobileNet-v2 \cite{sandler2018mobilenetv2} as the backbone, and then add a linear layer followed by a softmax layer. We set learning rate as 0.001, batch size as 128, and training epoch as 5. We split the synthetic dataset into 90\% training dataset and 10\% validation dataset, and we select the model with the highest validation accuracy during training as the final model. \\
\noindent\textbf{YOLOv8 Model.} For the non-popular object detection task, we use the pre-trained YOLO-v8 detection model \cite{Yolov8Ultralytics} as the specialized model. We set learning rate as 0.01, batch size as 16, and training epoch as 5. We also split the synthetic dataset into 90\% training dataset and 10\% validation dataset, and select the model with the highest validation accuracy. Note that when we generate the synthetic images with non-popular target objects (i.e. the pill bottles), we randomly place the target objects in the images and then use the fine-tuned background DM models to fill up the background. Therefore, we can directly obtain the labels of the target objects, i.e. the positions of target objects.

\subsection{Artifact}
The artifact for \ToolX~ is available at: \url{https://github.com/bitzj2015/SpinML-Artifact}.

\end{document}